%% file: ms.tex
\documentclass[12pt]{article}
\usepackage[top=1in,bottom=1in,left=1in,right=1in]{geometry}
\pdfoutput=1
\usepackage{times}

\usepackage{soul}
\usepackage{url}
\usepackage[small]{caption}
\usepackage{graphicx}
\usepackage{amsmath}
\usepackage{booktabs}
\usepackage{algorithm}
\usepackage{algorithmicx}
\usepackage{algpseudocode}
\usepackage{amssymb}
\usepackage{bbm,xspace,dsfont}
\usepackage{subcaption}
\usepackage{enumitem}
\usepackage{mathtools}
\newcommand*\samethanks[1][\value{footnote}]{\footnotemark[#1]}
\newcommand{\user}{\mathcal{U}}
\newcommand{\myitem}{\mathcal{I}}
\newcommand{\E}{\mathcal{E}}
\newcommand{\metric}{\mathcal{M}}
\newcommand{\graph}{\mathcal{G}}

\newcommand{\aucrelk}{\mbox{pAp}@k\xspace}

\newcommand{\medres}{MEDRES\xspace}

\newcommand{\modifiedGCN}{AL-GCN\xspace}

\newcommand{\gcnp}{\modifiedGCN}
\newcommand{\medresM}{\modifiedGCN-MEDRES\xspace}

\newcommand{\colf}{Collab\xspace}
\newcommand{\conf}{CoF\xspace}
\newcommand{\papk}{\textup{\textrm{pAp@k}}\xspace}
\newcommand{\preck}{\textup{\textrm{prec@k}}\xspace}

\newcommand{\pauc}{\textup{\textrm{pAUC}}\xspace}

\newcommand{\team}{\mathcal{T}}

\usepackage[colorinlistoftodos,textwidth=.4in]{todonotes}

\begin{document}

\title{Rich-Item Recommendations for Rich-Users: Exploiting Dynamic and Static Side Information}
\author{Amar Budhiraja\thanks{Equal Contribution.}\\
		Microsoft Research\\
	amar.budhiraja@microsoft.com
	\and
	Gaurush Hiranandani\samethanks\\
	UIUC\\
	gaurush2@illinois.edu
	\and
	\hspace*{25pt}Darshak Chhatbar \\
	\hspace*{25pt}Microsoft Research \\
	\hspace*{25pt}t-dachha@microsoft.com
	\and
	Aditya Sinha\\
	Microsoft Research\\
	t-adisi@microsoft.com
	\and
	\hspace*{25pt}Navya Yarrabelly\\
	\hspace*{25pt}Microsoft Research \\
	\hspace*{25pt}navya.yarrabelly@gmail.com
	\and
	Ayush Choure\\
	Microsoft Research\\
	aychoure@microsoft.com
	\and
	\hspace*{40pt}Oluwasanmi Koyejo\\
	\hspace*{40pt}UIUC \\
	\hspace*{40pt}sanmi@illinois.edu
	\and
	\hspace*{30pt}Prateek Jain\\
	\hspace*{30pt}Microsoft Research\\
	\hspace*{30pt}prajain@microsoft.com
}

\date{\today}

\flushbottom
\maketitle

\begin{abstract}
	In this paper, we study the problem of recommendation system where the users/items to be recommended are "rich" data structures with multiple entity types and with multiple sources of side-information in the form of graphs. We provide a general formulation for the problem that captures the complexities of modern real-world recommendations and generalizes many existing formulations. 
% In our formulation,  each user/document that requires a recommendation is modeled by a set of static entities and a dynamic component. Similarly, each item/tag to be recommended is also modeled by a set of static entities and a dynamic component. 
In our formulation,  each user/document that requires a recommendation and each item/tag that is to be recommended, both are modeled by a set of static entities and a dynamic component. 
The relationships between entities are captured by several weighted bipartite graphs. To effectively exploit these complex interactions and learn the recommendation model, we propose \medres -- a multiple graph-CNN based novel deep-learning architecture. \medres uses \gcnp, a novel graph convolution network block, that harnesses strong representative features from the underlying graphs. 
% , which effectively exploits complex interactions and learns the recommendation model. 
Moreover, in order to capture highly heterogeneous engagement of different users with the system and constraints on the number of items to be recommended, we propose a novel ranking metric \papk along with a method to optimize the metric directly. 
% Moreover, in order to capture highly heterogeneous engagement of different users with the system and constraints on the number of items to be recommended, we propose a novel ranking metric \papk. We also provide a method  that directly optimizes the proposed non-convex computationally hard \papk metric. 
We demonstrate effectiveness of our method on two benchmarks: a) citation data, b) Flickr data. In addition, we present {\em two} real-world case studies of our formulation and the \medres architecture. We show how our technique can be used to naturally model the message recommendation problem and the teams recommendation problem in the Microsoft Teams (MSTeams) product and demonstrate that it is 5-6\% points more accurate than the production-grade models.
\end{abstract}

\section{Introduction}\label{sec:intro}

\input{intro.tex}

%\section{Related Works}
%\input{related}

\section{Problem Formulation}
\input{prob}

\section{Method}
\input{method}
\section{Experimental Setup}\label{sec:exp}
\input{setup}

\section{Case Study 1: MS Teams Message Recommendation}
\input{exp_teams}

\section{Case Study 2: Team Recommendation (MS Teams)}
\input{exp_teamsrec}

\section{Benchmark Dataset Experiments}

\input{exp_citation}

%\section{Related Work}
%\input{related_work}

%\section{Case Study: Picture-Group Reco.}
%\input{exp_Flickr}

\section{Conclusion and Future Work}
\input{conclusion}

\bibliographystyle{plain}
\bibliography{ms}
\end{document}

%% file: intro.tex
Recommendation systems are a mainstay of most online systems and have been modeled using various approaches such as (inductive) matrix completion, nearest neighbor based predictions, content filtering ~\cite{koren2009matrix,jain2013provable,van2017graph}. %The emphasis on improving user experience in modern products via recommendations has led to multiple approaches in the Machine Learning (ML) community \cite{koren2009matrix,jain2013provable,van2017graph}. 
However, typical formulations simplify the entire system significantly, e.g., modeling users and items as a single static entity (collaborative filtering).% or manually design representations using the provided side-information in the form of a graph over users/items (content filtering). 
While such simple abstractions  provide tractable formulations that can be addressed using rigorous algorithms, they ignore key subtleties of the product environment. In a product system, information is available in multiple and varied forms, all of which may be important for the recommendation module to perform to its full potential.  To capture all the  information in the system and yet avoid the need for changing the model formulation with the addition of a new form of information, several practical recommender systems adhere to the \emph{content filtering} based approaches. In these approaches, all joint features for (user, item) tuples, extracted from the product, are fed into a classifier~\cite{li2013recommendation}. However, designing features for these approaches is a difficult problem, especially for real-world systems with several sources of information. Consequently, such methods suffer from issues such as poor accuracy and a difficult maintenance cycle.

Recent works in the domain of Heterogeneous Information Network (HIN) have proposed to better model such real-life systems with multiple heterogeneous sources of side-information, by constructing heterogeneous graphs and typically use certain metapaths side-information \cite{dong2017metapath2vec} or powerful attention/LSTM based models \cite{wang2019heterogeneous} to extract meaningful information from the heterogeneous graphs. However, these methods still, in general, model a user or item as fixed static entity with a few attributes, so do not capture challenges with many real-world systems where the users/items themselves can be "rich", i.e., are associated with multiple static and dynamic entities (see below). 

In this paper, we propose a general formulation for the recommendation problem with rich users and items. In particular, we model the real-world scenarios by considering two types of entities: (a) \emph{static} and (b) \emph{dynamic}. In a given time window, static entities are (mostly) fixed while dynamic entities are generated and modified at a high rate. For example, consider the citation prediction problem~\cite{yu2012citation}, where the goal is to predict the set of papers that should be cited by a given paper. Here, the authors and the conferences represent the static entities because their attributes/behavior is fixed in a considerably large time window. On the other hand,  the content of the paper (e.g. title/abstract) represents the dynamic entities, as they are generated at a considerably high frequency. 

% like the existence of multiple static entities in the system and multiple graphs describing the behavior of the entities. 
Given this notion of static and dynamic entities, we formalize both the \emph{user} -- an entity that seeks out the recommendation -- and the \emph{item}--an entity to be recommended --as a collection of static entities as well as a dynamic component. For example, consider an author who is seeking citation recommendations for her paper in preparation, which she plans to submit to a particular conference venue. Here, the user (i.e. the paper) is composed of static entities such as author and conference venue while it's dynamic component is the paper title or abstract. Similarly, the research papers to be recommended are composed of their authors (static), conference venues (static) where they were published, and their title or abstract (dynamic). We call this representation a `rich' representation of users and items. Notice that many items or users may have the same static entities but a different dynamic component.
% Past information in the form graphs between static entities such as how many times the author has cited a particular author may be exploited to provide citation recommendations. 
% For example, an author (user) seeking citation recommendations (items) for her research paper in preparation  
% a research paper is composed of static entities such as author and conference venue, whereas the paper title or abstract forms the dynamic component of the paper. We similarly formulate an item to be recommended as a collection of static entities and a dynamic component. 

Standard systems log a multitude of behavioral information about static entities using their past interaction with each other. These interactions, relationships, and engagements can be naturally represented using graphs. 
% co-author graph, where the edges denote the number of papers the authors have published together. Another example is 
One example is the author-conference graph, where an edge denotes the number of papers an author has published in a particular conference. Typical systems would have multiple such graphs among the entities. %Finally, we assume a labeled training dataset where for a given (user, item) pair, we have the ground-truth about how relevant the item is for the user. 

%for graphs, a lot of methods have been proposed. hetgcn, gat, ngm, gcn blah blah. but they don't combine properly. and they don't take broadviewpoint. 

Using the above mentioned "rich" representation for users/items along with their behavior captured by multiple graphs, we formulate the recommendation problem as that of finding relevance of an item for a user. %, where the user and the item are represented by multiple static entities and individual dynamic components. 
The training data consists of labeled (user, item) tuples and multiple bi-partite graphs between static entities. For this formulation, we propose a novel architecture, Multiple Entity based Deep REcommendation System (\medres), that combines the multiple entities and graphs associated with users and items via graph embedding techniques to embed each entity in a vector space after accumulating information from multiple graphs for that entity. We then 
% As our user/items are a collection of entities along with dynamic component, 
concatenate the entity embeddings with the dynamic component to obtain richer representations of users as wells as items. These representations are then fed to a multi-layer perceptron (MLP) to infer  how relevant an item is for a user. 
% The architecture is an end-to-end in the sense that it learns the embedding and infers the relevance simultaneously and can be trained using standard stochastic gradient descent based techniques. 
% Using standard stochastic gradient descent (SGD) based techniques, this end-to-end architecture learns multiple embeddings of multiple entities and infers the relevance, simultaneously. 
% \medres is an end-to-end architecture that can be trained for the required metric.

Now, in general, \medres can be combined with any standard graph embedding technique like Graph Convolution Networks (GCN) \cite{kipf2016semi}, Graph Recurrent Neural Networks (GRNN) \cite{you2018graphrnn}, Node2Vec \cite{grover2016node2vec} to process the given multiple graphs and produce representations for static entities. GCN, in particular, has been shown to be one of the most successful techniques in this domain and allows for efficient end-to-end training of the architecture. However, GCN is designed for semi-supervised settings where a node in the graph can correspond to only one labeled data point. So GCN's goal is to do label propagation over the graph. In our case, a node can correspond to multiple labeled points, and the total volume of labeled data is substantial, so the goal of graph-processing is to extract powerful features by training with the labeled data. To this end, we propose a novel \gcnp block, that builds upon GCN but adds residual connections, and also allows for learning non-linear functions of edge-weights to extract powerful features. We observe empirically that \gcnp indeed improves the performance of \medres significantly, and due to it's generality as a Graph Neural Network block, can be used in various other graph embedding applications. 

% to embed each entity in a vector space via combining information from multiple graphs for that entity. We then 
%% As our user/items are a collection of entities along with dynamic component, 
%concatenate the entity embeddings with the dynamic component to obtain richer representations of users as wells as items. These representations are then fed to a multi-layer perceptron (MLP) to infer  how relevant an item is for a user. 
%% The architecture is an end-to-end in the sense that it learns the embedding and infers the relevance simultaneously and can be trained using standard stochastic gradient descent based techniques. 
%% Using standard stochastic gradient descent (SGD) based techniques, this end-to-end architecture learns multiple embeddings of multiple entities and infers the relevance, simultaneously. 
%\medres is end-to-end architecture that can be trained for the required metric.%that we propose and finally care about.
%% , which is discussed as follows.
%
%% While designing accurate algorithms for a recommendation system is critical to the success of the system, evaluating the performance of the algorithm is an almost equally critical problem. %The reason is that standard metrics like AUC or NDCG metrics quite often fail to capture the key nuances of a recommendation system. 
% As mentioned above, we can optimize \medres for standard metrics like AUC, partial-AUC, Precision@k \cite{cortes2004auc,narasimhan2017support,le2007direct}.
One crucial advantage of \medres is that it is an end-to-end architecture that can be trained for the required metric. Thus, we may optimize \medres for standard metrics like AUC, partial-AUC (\pauc), Precision@k (\preck) \cite{cortes2004auc,narasimhan2017support,le2007direct}. However, existing metrics struggle with two well-known characteristics of real-world recommendation systems: a) the number of items ($k$) to be recommended per user is bounded, b) a large variance in the number of positives across users (or nodes in the graph). Due to a combination of  the above-mentioned issues,  existing metrics  can fail to capture the "correct" solution and can paint a misleading picture (see Section~\ref{sec:papk}). In this paper, we design a new metric --partial-AUC+precision@k (\aucrelk)-- that combines \pauc and \preck, and is aware of the above-mentioned characteristics. In fact, the metric is used in {\em multiple production systems} deployed by MSTeams. In this paper, we formalize the metric and provide an optimization algorithm for the same that we use to train our \medres architecture. %, and attempts to capture both the fact that number of positive labelled points per user can have high variance and the number of items to be recommended is limited. 

Finally, we study the wide-applicability of our formulation and effectiveness of our algorithm by modeling two problems in MSTeams: message recommendation and Team/Group recommendation. We observe that \medres provides impressive gains over the production models for the message recommendation system deployed in MSTeams. 
We also apply our \medres architecture to  two publicly available datasets: a) citation dataset, b) Flickr dataset. For both these datasets, we show how challenging recommendation problems can easily be instantiated in our framework, and can provide up to $5-6\%$ more accurate recommendations than standard baselines.

%describe a natural recommendation problem, and benchmark our algorithm against two standard and state-of-the-art techniques: a) collaborative filtering via tensor completion, b) content filtering with Gradient Boosted Decision Tree (GBDT) and Neural Network (NN) classifier. We demonstrate that our algorithm that does not require any hand-designed features can be significantly more accurate than either of these two approaches. 

% Finally, we present a case study of message recommendation on Microsoft Teams (MSTeams) which is a real-world recommendation system. MSTeams is a fast-growing product with several million users but can also expose its users to a significant information overload due to a large number of incoming messages for each person. We discuss various subtleties in this problem as well as how \medres can be applied effectively. Particularly, we observe that \medres achieves 5-6\% improvement in performance (as measured by \aucrelk) over production grade models. 
% % \vspace{-1mm}
In summary, following are the key contributions of the paper: 
\begin{itemize}[noitemsep, leftmargin=2em]
	\item We propose a new formulation for the modern recommendation systems via "rich" users and items, that effectively captures dynamically generated content, static entities, and multiple relationships among the entities (Section~\ref{sec:prob}). 
	\item We propose \medres, a multiple Graph-CNN based architecture to exploit the above formulation (Section~\ref{sec:method}). \medres' backbone architecture  critically uses \gcnp, a novel graph embedding technique. 
	\item We propose a new metric \aucrelk, that captures subtleties in several real-world recommendation systems. We also provide an algorithm to optimize the proposed metric (Section~\ref{sec:method}). The \aucrelk metric is used for evaluation in multiple real-world recommendation systems. 
	\item We present empirical validation of our proposed solutions on two benchmark datasets. We also demonstrate our technique on two real-world recommendation systems for MSTeams\footnote{MSTeams is an enterprise chat+meeting product with millions of active users} where our method significantly outperforms strong production-grade models, and is being deployed in the product. %The two systems are: a) message recommendation system, b) Teams recommendation system in MSTeams--an enterprise product with millions of active users--. %We compare our method against a production-grade model developed over 1-2 years and show that our method is able to significantly outperform it after a few hours' worth of training (Section~\ref{sec:exps}). 
\end{itemize}

\subsection{Related Works}
Heterogeneous Information Networks (HINs) is a popular approach to model multiple heterogeneous entities in the system, and are used in several representation learning problems \cite{yu2013recommendation,zhao2017meta,fu2017hin2vec}. The core idea of HINs is the emphasis on the graph's heterogeneous entities, but the user-item pairs are considered as static nodes in the graph, thus do not allow for rich users/items with multiple static entities or dynamic component. For example, in the formulation of HINs for citation networks, if citations are required for a new paper, ideally the entire model needs retraining. Our formulation, on the other hand, allows for a more flexible and richer aspect of user-item pairs by allowing items to be dynamic and users and related entities to be static in the system. 
For instance, in the case of a new paper, we model it as a dynamic item represented by the paper's content vector in our formulation. 
% Thus, for the previous example of citation networks, we model a new paper as a dynamic item represented by the paper's content vector in our formulation. 
In another related study\cite{sunmulti}, the authors also use GCNs to learn embeddings from multiple graphs to compute relevance between user-item pairs, but their approach is also restricted to a static set of items, contrary to the proposed framework which can model items dynamically. Thus, HIN and related works are complementary to our approach, where their architecture can be used as the graph embeddings extraction block in our novel framework.

%Heterogeneous Information Networks: there are several works that cpature heterogeneous entities, but the user, item pairs are static nodes in this graph. we allow for more rich pairs.

Graph Representation Learning (GRL) has been widely studied in the literature, and several GRL approaches have been proposed such as Graph Convolutional Networks (GCN) \cite{kipf2016semi}, Graph Recurrent Neural Networks (GRNN) \cite{you2018graphrnn}, node2Vec \cite{grover2016node2vec}, DeepWalk \cite{deepwalk}, NGM \cite{bui2018neural}, etc. but all these methods take graph weights as given, which could be problematic because of noise in the real world graphs. GAM \cite{stretcu2019graph} is an exception to this, but the method proposed in GAM still uses the same weighing scheme, with augmentation of the propagated weights. \modifiedGCN on the other hand learns the graph weights in an end-to-end fashion, thereby extracting task-specific signals from the graph. %Our method is different from collaborative filtering and content filtering techniques since it is hard and time consuming process to design these features, and deep learning methods, like \medres have proven to be immensely successful in automatic feature learning directly from the raw data.

Finally, there exists a vast literature on metrics in recommendation system. AUC, \pauc, \preck, and NDCG@k are the most well-known and popular metrics in this space~\cite{narasimhan2017support, kar2015surrogate,valizadegan2009learning, liu2009learning, clemencccon2009empirical}. However, these metrics do not address the bipartite ranking problem with varied user engagement levels and limitations on the number of recommended items, a problem at the heart of several modern recommendation systems. The \papk metric is designed to address the above-mentioned problem and is in use by multiple production systems at MS Teams (see Section~\ref{sec:method}). %In this paper, we also propose a metric, pAp@k, which aims at capturing the subtle details of evaluation of recommendation systems, which are currently not addressed by existing metrics. AUC, which is a widely accepted metric for example, computes a holistic metric and can paint a wrong picture about the solution. pAUC \cite{narasimhan2017support} was proposed to resolve some of these issues with AUC but since it still considers the whole set of positives and partial set of negatives, it can express a wrong sense of performance. These issues with AUC and pAUC are further exacerbated when different users contribute to different volumes of data points. Precision@k (prec@k) fails to capture rankings within the top-k items, and a ranking function which produces high quality results at the top of k items is ranked equal to a function which has produced the same number of ones but those ones are at the bottom in the top-k. NDCG@k, on the other hand, focuses on solely on the ranking of the top-k items and penalizes relevant results heavily even if they occur near the top, which is may paint a pessimistic picture in case of online recommendations. The proposed metric, \aucrelk aims at addressing these issues, which we discuss at length in Section XXX.

%GCNs: several graph embeddings techniques like gcn, grnn, node2vec, deepwalk,ngm etc. but they take graph weights as it is, which is problematic. ngm is an exception but their graph weighting is still same, they just augment wiht propogated weights. GCNP more powerful as we are in relatively data rich setting. 

%Collaboarative and content filtering: hard to design features for

%Metrics: several works like ndcg@k, precision@k, auc, auc@k, pauc, but none capture the issues mentioned in the intro. 

%% file: prob.tex
\label{sec:prob}

Given a user/document $\user$ and an item $\myitem$, the goal is to find relevance score of $\myitem$ for $\user$. Both  $\user$  and $\myitem$ are complex entities with several static sub-entities and a dynamic component. Let $\E^{(1)}, \dots, \E^{(E)}$ be given {\em entity types} for some $E\in \mathbb{Z}_+$. An entity type can be a set of authors, groups, venues etc. $\E^{(i)}_\user$ denotes the $i$-th entity type for user. For example, if $\E^{(1)}$ denotes the set of groups, then $\E^{(1)}_\user$ denotes the group of $\user$. With these notations in place, $\user$ is now defined as: \begin{equation}\label{eq:user}\user=(\E^{(a_1)}_\user, \dots, \E^{(a_{k_1})}_\user, \zeta(\user)),\end{equation}
where each $\user$ is associated with $k_1$ entity types: $\E^{(a_i)},\ i\in [k_1]$, and  $[k] = \{1,\dots,k\}$ denotes the index set for $k \in \mathbb{Z}_+$. %$\E^{(a_{k_1})}_\user$ denotes the entity instance of type $\E^{(a_{k_1})}$ corresponding to user/document $\user$. 
$\zeta(\user)$ is the {\em dynamic} part of $\user$. 
For example, for a research paper $\user$ being written for conference $C$ by author $A$, the entities are conference $\E^{conf.}_\user = C$ and author $\E^{auth.}_\user = A$. 
The {\em dynamic} part is the text of the paper, i.e., $\zeta(\user)\in \mathbb{R}^{D_U}$ can be say $D_U$-dimensional word2vec embedding~\cite{mikolov2013distributed} of the paper's abstract and title.  Similarly, an item $\myitem$ is defined as:
 \begin{equation}\label{eq:item}\myitem=(\E^{(b_1)}_\myitem, \dots, \E^{(b_{k_2})}_\myitem, \nu(\myitem)),\end{equation}
 where $\nu(\myitem)\in \mathbb{R}^{D_I}$ denotes the {\em dynamic} part of the item with $D_I$ being the dimension of the dynamic component. For instance, a citation item has two static entities (authors and conference venue) and it's dynamic part is the word-embedding of paper title. The goal is to recommend citations i.e. a set of items for a new research paper $\user$. 
 
Additionally, a multitude of behavioral information about static entities can be collected using  their past  interactions. These interactions are expressed via (multiple) bipartite graphs. For example, we may have an  author-conference graph where an edge represents the number of times an author has published at a conference. 
We denote these bipartite graphs between entities $\E^{(a)}$, $\E^{(b)}$ as: % \begin{equation}\label{eq:graph}
$ G^{a,b}=(V^{a,b}, A^{a,b}), $ 
%\end{equation}
where $V^{a,b}$ is the set of nodes and $A^{a,b} \in \mathbb{R}^{|\E^{(a)}| \times |\E^{(b)}|}$ is the adjacency matrix of the graph. The rows and columns in $A^{a,b}$ are associated with entity instances of type $\E^{(a)}$ and $\E^{(b)}$, respectively. For any two entity types $\E^{(a)}$ and $\E^{(b)}$, we may have multiple bipartite graphs denoting different interactions. Thus, let $\graph^{a,b}=\{G^{a,b,1}, \dots, G^{a, b,|\graph^{a,b}|}\}$ be the set of graphs between entities $\E^{(a)}$ and $\E^{(b)}$. Furthermore, let the set of all graphs be represented by $\graph=\{G^1, \dots, G^{|G|}\}$. 
In addition to such graphs, we are provided the following dataset for training: \begin{equation}\label{eq:dataset}\mathcal{D}=\{(\user_1, \myitem_1, y_1), \dots,(\user_n, \myitem_n, y_n)\},\end{equation} 
where $y_i \in \{0, 1\}$ is the label of the $i$-th (user, item) pair and denotes the binary relevance of item $\myitem_i$ for user $\user_i$. Given all this information, the goal is to find a scoring function that computes relevance of $\myitem$ for $\user$. %optimizes certain metric $\metric$ given user $\user$, item $\myitem$ and entity graphs $\graph$. 
That is, given $\mathcal{D}$, entities $\E^{(1)}, \dots, \E^{(E)}$, graphs $\graph$, the goal is to find a scoring function $s(\user, \myitem)$ that works best for a certain metric $\metric(\mathcal{D})$. For example, for binary labels $y_i$'s, the goal would be to find a score $s$ s.t.  $s(\user_i, \myitem_i)>s(\user_j, \myitem_j)$ if $y_i>y_j$. 

Note that most of the existing recommender system formulations can be expressed as a special case of this general formulation. For example, typical collaborative filtering \cite{koren2009matrix} is defined by $\user=(\E^{user})$, $\myitem=(\E^{item})$. That is, there are only two entity types: the set of users and set of items. Furthermore, users/items do not have any dynamic content. Berg et al.~\cite{van2017graph} extended this model by considering the ratings graph $G^{\user,\myitem}$ between users and items. That is, edge $(\user,\myitem)$ exists in graph if $\user$ has given $\myitem$ a fixed rating. This model is also a special case of our formulation where we have only two types of entities $\user, \myitem$, and the only graphs are the ratings graphs. 

Similarly, inductive matrix factorization \cite{jain2013provable} models the problem as: $\user=(\zeta(\user))$, $\myitem=(\E^{item})$, where the set of entities contains only the item entity itself. $\zeta(\user)$ is the dynamic user/document feature of $\user$. Recently, several papers~\cite{rao2015collaborative,monti2017geometric, zhao2017meta,zhang2019heterogeneous,wang2019heterogeneous} have studied the matrix completion problem with graph based side-information and several attributes. Typically, these works model the users/items as: $\user=(\E^{user})$, $\myitem=(\E^{item})$ with graphs among users, items, and their attributes. 
Finally, the standard content filtering approaches~\cite{ahn2008new,kim2011collaborative,li2013recommendation} do not exploit graph structure and directly models the $(\user, \myitem)$ pair as: $\phi(\user,\myitem)$, i.e., via hand-coded features for $\user,\myitem$ pair.  This shows that most of the existing recommendation problems are a special case of our general framework. Moreover, unlike standard formulations, our general framework leads to a real-world recommendation  system that captures much more complex entities and stitches together various sources of information.

%% file: method.tex
\label{sec:method}

We first discuss the proposed architecture \medres, which stitches several types of interactions amongst multiple entities to recommend $\myitem$s for $\user$s followed by \modifiedGCN which extends GCN by using non-linear adjacency matrix. We then discuss the new metric, $\aucrelk$, which appropriately measures the effectiveness of a recommendation system in a classification+ranking scenario such as ours followed by an optimization method for the proposed metric. 
% Lastly, we discuss how to infer using the resulting end-to-end architecture.
Lastly, we discuss the training of MEDRES for optimizing \papk metric.

\subsection{\medres Architecture}
\input{architecture}

\subsection{Adjacency Learnt GCN (\modifiedGCN)}
\input{NOA-GCN}

\subsection{The Performance Metric -- \aucrelk}
\input{pap-at-k}
\subsection{Training: optimizing \papk}
\input{infer}

%% file: architecture.tex
In this section, we propose \medres, an end-to-end architecture, to learn multiple embeddings for all entity instances of every entity type. We then concatenate all these embeddings along with dynamic content to obtain rich $[\user, \myitem]$ representations and feed them into a multi-layer perceptron (MLP) network to obtain the final relevance score. 

\begin{figure}[t]
\begin{subfigure}{.48\textwidth}
\includegraphics[scale=0.23]{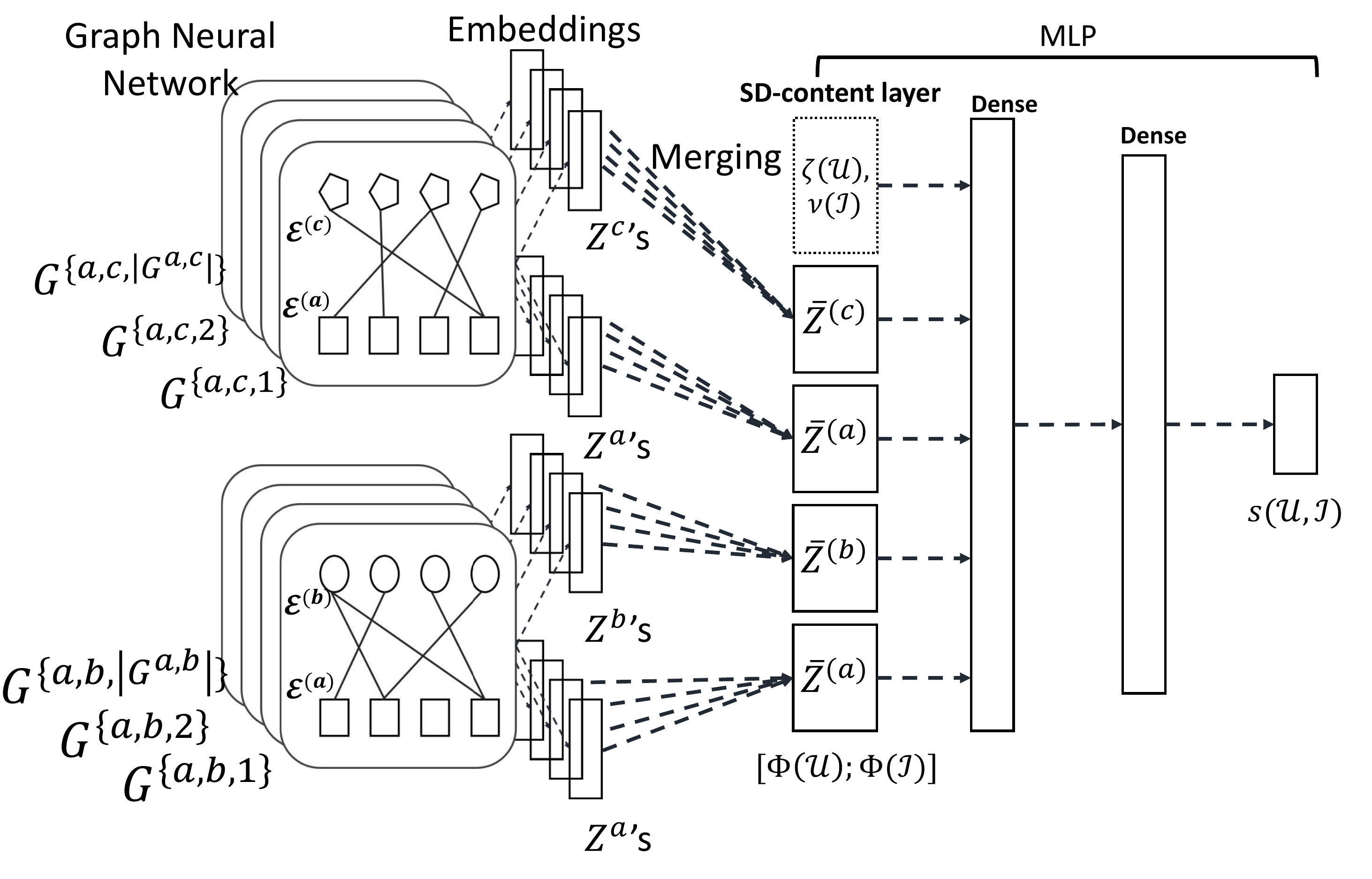}
  \caption{\medres: Embeddings are generated using any graph neural network technique. These embeddings are merged together to obtain multiple \emph{Emb-merge} embeddings. Dynamic content is augmented with all the \emph{Emb-merge} embeddings in the \emph{S-D Content} layer, which is then passed to an MLP for binary classification.}
  \label{fig:arch}
\end{subfigure}\hspace*{5pt}
\begin{subfigure}{.48\textwidth}
	\includegraphics[scale=0.25]{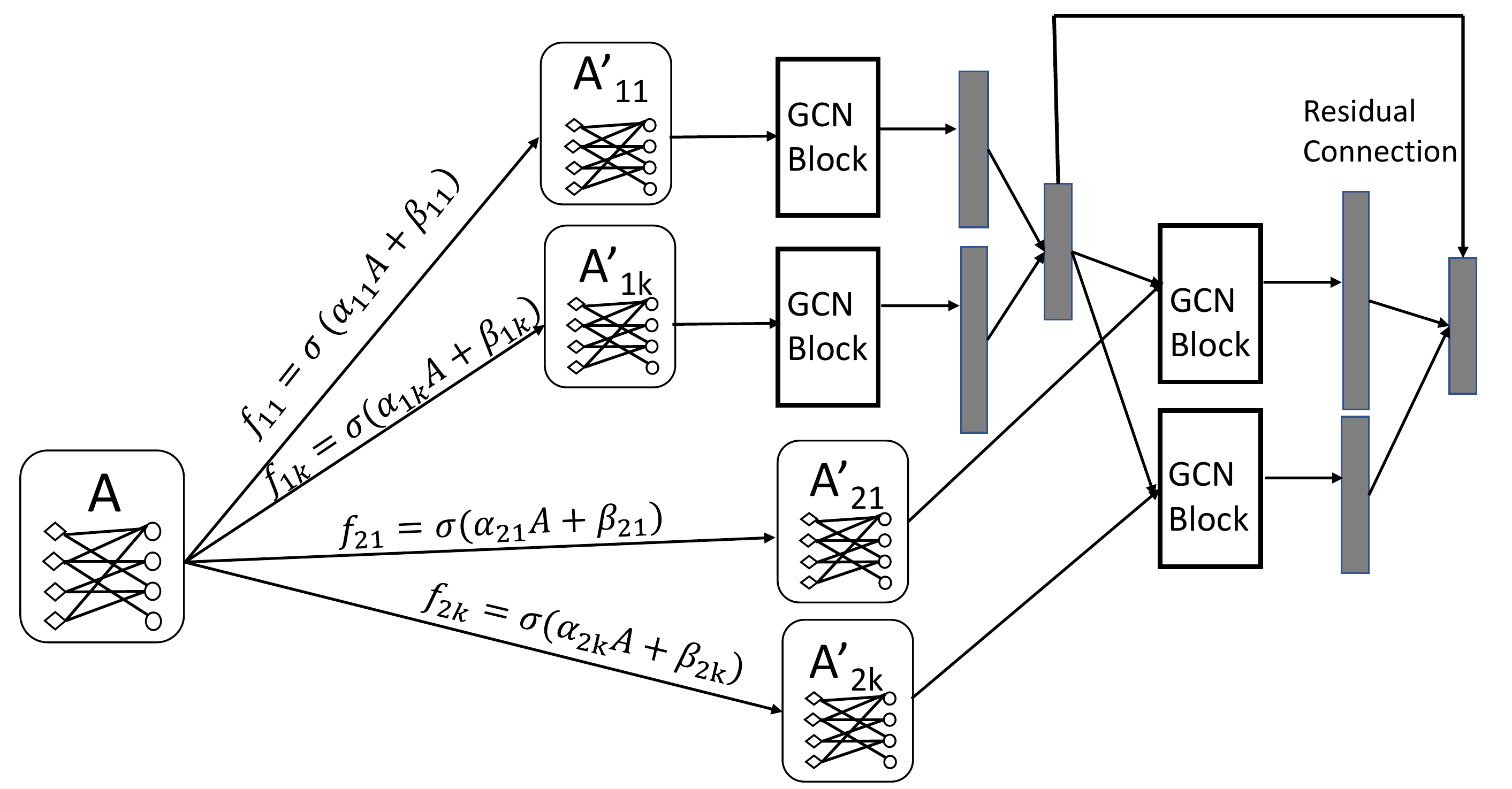}
	\vskip 0.6cm
	\caption{\modifiedGCN: First, the adjacency matrix is transformed by applying k functions at each layer, then these transformed matrices are used by the GCN Block to propagate embeddings, and then results are aggregated at the output of each layer. In the end, residual connections are established between outputs of all layers. 
% 	\gh{can explain what each component is and equations. Will also make caption length compatible to the neighboring figure. Also, align the captions  of tables and figures using vskip/vspace. Will improve paper aesthetics. Also, can align shapes better so to make figure aesthetics better.}
	}
	\label{fig:noa-gcn}
\end{subfigure}
\caption{}
\end{figure}

Unlike existing methods~\cite{li2013recommendation}, which use feature extraction from graphs and a separate classification model, we learn all the parameters simultaneously for training our end-to-end architecture while optimizing the proposed metric $\aucrelk$. Please see Figure~\ref{fig:arch} for a block representation of the framework. 

\label{ssec:arch}

Let us suppose that we are given a bipartite graph $G^{a,b}$ between two entities $\E^{(a)}$ and $\E^{(b)}$. For simplicity of exposition, let us represent the graph as $G = \{V^{(a)}\cup V^{(b)}, A\}$, where $V^{(a)}$ and $V^{(b)}$ correspond to entity instances of entity types $\E^{(a)}$ and $\E^{(b)}$, respectively, and $A$ is the adjacency matrix between the nodes defined as:
$$
A = 
\begin{pmatrix}
  \begin{matrix}
  0
  \end{matrix}
  &  A^{a,b} \\
  (A^{a,b})^T  &
  \begin{matrix}
  0
  \end{matrix}
\end{pmatrix}.
$$
Let $F^{(a)}\in \mathbb{R}^{|V^{(a)}|\times p}$ and $F^{(b)}\in \mathbb{R}^{|V^{(b)}|\times q}$ be the feature matrices of entity instances associated with $\E^{(a)}$ and $\E^{(b)}$, respectively\footnote{In the absence of feature vectors, we can select the one-hot encoding of nodes for $\E^{(a)}$.}. Then, the embeddings for $a$ and $b$ are computed as: % $f^{(a)}_i$ being the $i$-th row of $F^{(a)}$ and $f^{(b)}_i$ being the $i$-th row of $F^{(b)}$.
% using formulation in \cite{kipf2016semi}, we have:
% Now, using the GCN technique from \cite{kipf2016semi}, we define: 
% \begin{equation}Z = \sigma({D}^{-\frac{1}{2}}(A+\lambda I){D}^{-\frac{1}{2}}B\cdot W),\end{equation}
{\footnotesize 
\begin{equation}
\begin{bmatrix}
%   \begin{matrix}
  Z^{(a)} \\
  Z^{(b)}
%   \end{matrix}
\end{bmatrix}
  = f_{e}(A, F) %\Big((f(\tilde L) + I)FW),
\label{eq:approx}
\end{equation}}
where $f_{e}$ is an embedding function (such as  Graph Convolution Networks~\cite{kipf2016semi}) that takes as input the adjacency matrix $A$ and the feature matrix $F$ of $G^{a,b}$ and outputs  $Z^{(a)}\in \mathbb{R}^{|V^{(a)}|\times D_{{\tiny \mbox{CN}}}}$, $Z^{(b)}\in \mathbb{R}^{|V^{(b)}|\times D_{{\tiny \mbox{CN}}}}$, the embeddings of entities $\E^{(a)}$ and $\E^{(b)}$. %generated from the bipartite graph, $G^{a,b}$. $f_{e}$ takes as input the adjacency matrix $A$ and the feature matrix $F$ of $G^{a,b}$, where 
Here, $F$ is given by: $F=[F^{(a)}\ \ 0; 0\ \ F^{(b)}]$. 
%{\footnotesize
%% \begin{equation}
%% F=
%%   \begin{bmatrix}
%%   F^{(a)} & 0 \\
%%   0 & F^{(b)}
%%   \end{bmatrix}W,
%%   \label{eq:gcp2}
%% \end{equation}
%\begin{equation}
%F=
%  \begin{bmatrix}
%  F^{(a)} & 0 \\
%  0 & F^{(b)}
%  \end{bmatrix},
%  \label{eq:gcp2}
%\end{equation}
%}

%In the architecture, we call this layer as \emph{GC-P2} layer and the corresponding embeddings such as $Z^{(a)}$ as \emph{GC-P2} embeddings (see Figure~\ref{fig:arch}). 

Now, as we discussed in Section~\ref{sec:prob}, there can be multiple graphs between entity types $\E^{(a)}$ and $\E^{(b)}$. For example, interactions such as `number of papers published' and `number of papers cited' in a conference by an author may be represented as two different graphs.
% one graph may represent the number of times an author has published in a particular conference,  may be represented by one graph and how many times author has cited a paper from a particular conference may denote another graph. 
We merge all the embeddings of an entity type learnt by $f_{e}$ independently from all the interaction graphs using a fully connected layer. We refer to this layer as the \emph{Emb-merge layer}. Formally, let  $\graph^{a,b}$ be the set of graphs between entities $\E^{(a)}$ and $\E^{(b)}$. Then, we first concatenate all embeddings i.e. $Z^{(a)}$'s~\eqref{eq:approx} of entity instances in $\E^{(a)}$ emerging from the graphs in $\mathcal{G}^{a,b}$ and merge them using a fully connected layer. The same is done for entity instances in $\E^{(b)}$ as described in the following formulation:
{\small 
\begin{equation}
\bar Z^{a,b} = 
\begin{bmatrix}
%   \begin{matrix}
  \bar Z^{(a)} \\
  \bar Z^{(b)}
%   \end{matrix}
\end{bmatrix}
  = RELU \left(
  \begin{bmatrix}
 Z^{(a)}_{1} \cdots Z^{(a)}_{|\graph^{a,b}|} \\
 Z^{(b)}_{1} \cdots Z^{(b)}_{|\graph^{a,b}|}
  \end{bmatrix}
  \begin{bmatrix}
  W^{(a)} \\
  W^{(b)}
  \end{bmatrix} + 
  \begin{bmatrix}
  c^{(a)} \\
  c^{(b)}
  \end{bmatrix}\right),
  \label{eq:embmerge}
\end{equation}}
where $\bar Z^{a,b} \in \mathbb{R}^{(|V^a| + |V^b|) \times D_{{\tiny \mbox{MN}}}}$ with $D_{{\tiny \mbox{MN}}}$ being the  \emph{Emb-merge} layer's embedding dimensions, $c^{(a)}, c^{(b)} \in \mathbb{R}^{D_{{\tiny \mbox{MN}}}}$ and $W^{(a)}, W^{(b)} \in \mathbb{R}^{\left(|\mathcal{G}^{a,b}|\ast D_{{\tiny \mbox{CN}}}\right) \times D_{{\tiny \mbox{MN}}}}$ are the biases and weights of the \emph{Emb-merge} layer, respectively.
% corresponding to the entities types $\E^{(a)}$ and $\E^{(b)}$. 
% We call this layer \emph{Emb-Merge} layer. 
Notice that we always have the flexibility to choose different dimensions $D_{{\tiny \mbox{MN}}}$ for different entity types. $\bar Z^{(a)}$ represents an embedding of entity instances in $\E^{(a)}$ with regards to $\E^{(b)}$. For example, at this stage, we have embedding of authors due to all the interactions with conferences.

After utilizing all the static information (graphs) to get entity embeddings, now we augment the dynamic content as features. That is, we represent every user/document $\user$ as: 
% \begin{equation}\Phi(\user)=[\bar Z^{1,2}_\user, \bar Z^{1,3}_\user, \dots, \bar Z^{E, E}_\user, \zeta(\user)]\in \mathbb{R}^{\hat{D}_1},\label{eq:fin_emb}\end{equation}
\begin{equation}\Phi(\user)=[\bar Z^{1,2}_\user, \bar Z^{1,3}_\user, \dots, \bar Z^{E, E}_\user, \zeta(\user)],\label{eq:fin_emb}\end{equation}
where $Z^{i,j}_\user$ is the \emph{Emb-merge} embedding of $\user$ from all the graphs between entities $\E^{(i)}$ and $\E^{(j)}$. If $\user$ is not of type $\E^{(i)}$ and $\E^{(j)}$, then $Z^{i, j}_\user=0$. We represent each item $\myitem$ similarly. 

% For instance, there can be user-author graphs and user-conference graphs, but there may not be author-conference graphs. In that case, the \emph{Emb-merge} layer for authors will be realized only from user-authors graphs, and it will be zero for author-conference graphs. Similarly, we represent each item $\myitem$ as: 
%\begin{equation}\Phi(\myitem)=[\bar Z^{1,1}_\myitem, \bar Z^{1,2}_\myitem \dots, \bar Z^{E, E}_\myitem, \nu(\myitem)]\in \mathbb{R}^{\hat{D}_2}.\end{equation} 
We call the concatenation of user-item representations $[\Phi(\user);\Phi(\myitem)]$ as the \emph{S-D-content} embedding, where S-D stands for `static'-`dynamic'. We then apply an MLP layer on the final \emph{S-D-content} as follows: 
\begin{equation}
s(\user,\myitem)=\sigma(RELU(RELU( [\Phi(\user);\Phi(\myitem)]M_1)M_2)M_3),
\end{equation}
where $M_i$'s are the weight matrices and $\sigma$ is the sigmoid function (for binary classification problem).  

%We further add \textbf{residual connections} \medres to make it more robust to overfitting and to allow richer embeddings of the entities for the FC layers for classification. To add the residual connections, we use the output of the \modifiedGCN filter at each hop, and concatenate the outputs together. Adding these residual connections allows the network to effectively encode information from multiple hops of a node's neighborhood, where each hop can play an individual role in the classification task. Through our experiments, we observed that adding these residual connections from multiple hops of \modifiedGCN always outperforms when taking embeddings from only the last hop of the architecture. 

%In the experiments (Section~\ref{sec:exps}), we consider two versions of \medres. One with polynomial-2 approximation  for obtaining the \emph{GC-P2} embeddings. We call it \medrest. Another version, referred as \medreso, uses polynomial-1 approximation (i.e. considering only the first term in the \emph{r.h.s} of~\eqref{eq:approx}) to obtain \emph{GC-P1} embeddings. 
% This variation is inspired from Berg et al~\cite{van2017graph}, where the authors have used polynomial-1 approximations for a single bipartite graph. 
%We empirically observe  that \medrest captures better `collaborative filtering' aspect than \medreso in two out of three datasets. 

%% file: NOA-GCN.tex
%\begin{figure}
%\includegraphics[scale=0.31]{figures/NOA.pdf}
%  \caption{\modifiedGCN: Adjacency matrix is transformed by applying k functions at each layer and results are concatenated at the end of each layer. At the end, output of all layers are concatenated to make residual connections.}
%  \label{fig:noa-gcn}
%\end{figure}

% \gh{I believe here we should remind the reader that some blocks in the previous section can be GCN or other models. But we propose something better -- like the same written in intro.}
\medres can be combined with any standard graph embedding technique like GCN, GRNN, Node2Vec to process multiple graphs, but now we propose \modifiedGCN, a new graph embedding technique which performs better than these techniques, when combined with \medres.
\modifiedGCN is a novel GCN style graph embedding technique that provides more powerful representations than the standard GCN methods (Section~\ref{sec:teamreco}). Now, a typical GCN layer is defined as: \begin{equation}
    \label{eq:gcn_layer} X^{l+1} = \sigma(AX^{l}W^{l}),
\end{equation}
where $X^{l}$ is $l^{th}$ layer embedding of GCN, $W^{l}$ is $l^{th}$ layer weight matrix, A is the Adjacency Matrix, and $\sigma$ is a non-linear function.
Let us consider a multiple-graph setting, where graphs $A^1...A^m$ are given between entities $\E^{(a)}$ and $\E^{(b)}$. In \modifiedGCN, we  apply a function on the adjacency matrix before using it for propagating embeddings over the graph. That is, \modifiedGCN requires functions $f_p:\mathbb{R}\rightarrow \mathbb{R}$ that are applied entrywise and acts on the aggregation of all graphs, i.e, $\hat{A}_{ij} = [A^1_{ij},A^2_{ij} \dots A^m_{ij}]$: 
\begin{equation}
    \label{eq:noa-gcn_layer} \Tilde{X}^{l+1} = [\sigma(f_1(\hat{A})\Tilde{X}^{l}W^{l}), \dots, \sigma(f_k(\hat{A})\Tilde{X}^{l}W^{l})],\ \ {\text where}\ f_p(\hat{A}_{ij}) = \sigma(\alpha_p\hat{A}_{ij} + \beta_p), 
\end{equation}
 $\Tilde{X}^{l} $ is the $l^{th}$ layer embedding, $W^{l} $ is the weight matrix, $\sigma $ is a non-linear function, $\alpha_p\in \mathbb{R}^m$ and $\beta_p\in \mathbb{R}$. Note that, $W^{\ell}$, $\alpha_p$, $\beta_p$ are all learnable parameters in this model.

% The function $f$ takes edge-weight $A_{ij}$ of the adjacency matrix and transforms it as defined below:
%\begin{equation}
%    \label{eq:noa-gcn-fn} f(A_{ij}) = \sigma(\alpha A_{ij} + \beta)
%\end{equation}
%where $\alpha$ is scalar learnable weight multiplied to all the edge-weights and $\beta$ is the bias added to it. Here $f(A_{ij}) ~ \epsilon ~ R^k $as we apply k such functions at each layer and hence we get k graphs. We concatenate result of Eq~\ref{eq:noa-gcn_layer} obtained from k functions at each layer. To take input multiple graphs, function $f$ can be modified as:
%\begin{equation}
%    \label{eq:multiple-graphs} f(\hat{A}_{ij}) = \sigma(W\hat{A}_{ij} + b)
%\end{equation}
%where $\hat{A}_{ij} = [A^1_{ij},A^2_{ij} \dots A^m_{ij}]$, W $~\epsilon~R^{mxk} $, b$~\epsilon~R^{mxk}$, where m is number of graphs and k is number of functions applied,  and $\sigma$ is a non-linear function like ReLU.

We further add \textbf{residual connections} in \medres to make it more robust to overfitting and to allow richer embeddings of the entities to the MLP layers for classification. To add the residual connections, we use the output of the \modifiedGCN filter at each hop and concatenate the outputs together. After concatenation, the final embeddings looks like:
\begin{equation}
    \label{eq:residual} Z = \Tilde{X}^1 \oplus \Tilde{X}^2 \oplus \dots \oplus \Tilde{X}^L,
\end{equation}
where Z is the final embeddings obtained from \modifiedGCN, $\Tilde{X}^l$ is the output of $l^{th}$ layer \modifiedGCN as defined in Eq~\eqref{eq:noa-gcn_layer} and $\oplus$ makes resiudal connections between output of these layers. Adding these residual connections allows the network to effectively encode information from multiple hops of a node's neighborhood. %, where each hop can play an individual role in the classification task. 
Through our experiments, we observe that adding these residual connections from multiple hops of \modifiedGCN always outperforms taking embeddings from only the last hop of the architecture. 
See Fig~\ref{fig:noa-gcn} for a visual illustration of the \modifiedGCN method. % where multiple Non-linear transformations are  applied to Adjacency Matrix and then forward pass of GCN is computed on this transformed Graph for each of them and then concatenated at the end. %Similar computation is done at all layers of \modifiedGCN and then output of all layers are concatenated to establish residual connections. Fig~\ref{fig:noa-gcn} shows only one graph but can be applied to multiple graphs in a similar way as described in Eq~\ref{eq:multiple-graphs}.

%% file: pap-at-k.tex
\label{sec:papk}
An important component of any strong recommendation system is the metric that is used to evaluate different methods, and it should capture and reflect the key nuances of the system. Modern systems typically use \preck, \pauc, AUC, NDCG@k style metrics for this purpose~\cite{cortes2004auc, narasimhan2017support, kar2015surrogate,valizadegan2009learning, liu2009learning, le2007direct}. 

Now, metrics like NDCG@k do not apply to bipartite ranking style problems where the goal is to rank positives above negatives, and gain/loss for ranking a positive above a negative should not change by their position in the list. For bipartite ranking, pAUC and \preck style measures are the most popular~\cite{cortes2004auc, kar2015surrogate}. However, both of them struggle when two key characteristics common to several recommendation systems are in effect jointly: (a) amount of engagement, i.e., fraction of positives, per user can have high variance, and (b) number of items to be recommended are limited,  

For example, let  $k=3$ and let there be $8$ points with $2$ positives. Let there be two methods that produce rankings $r_1$ and $r_2$ of the given points, and let $f_1$ and $f_2$ be the true labels corresponding to the ranked order. For example, let $f_1: \{0,0,1,1,0,0,0,0\}$ and $f_2: \{1,0,0,1,0,0,0,0\}$. The value for \preck is 0.33 for both $f_1$ and $f_2$, but it is clear that $f_2$ is a better ordering as a positive is at the top of the list in that case. Now, for a similar example with $k=3$ and $3$ positives, let certain methods produce ranked orders $f_3$ and $f_4$ where $f_3=\{1,1,0,0,0,0,0,1\}$ and $f_4=\{0,1,1,1,0,0,0,0\}$. Here for $k=3$, $f_3$ is clearly better but \pauc for threshold $3/6$ is $0.66$ for both the rankings. 

Naturally, the issue is that in the first case, a classification metric like \preck does not care for the ranking, while in the second case \pauc is not much aware of top-$k$ recommendations. 

Next, we present \papk, a novel metric which attempts to mitigate the above mentioned concerns with the existing methods. We first define \papk assuming only one user for which we have a given set of labeled points $(x_i, y_i), \cdots ,(x_n, y_n)$, where $x_i \in \mathcal{X}$ and $y_i \in \{0, 1\}$ and a scoring function $s:\mathcal{X}\rightarrow \mathbf{R}$. Let $n_+$ be the number of positives (i.e. $y_i=1$) in the  data and $n_-$ be the set of negatives ($y_i=0$). Let $S^+$ be the set of top-$\beta$ positives ordered by scoring function $s$, where $\beta=\min(n_+,k)$ and $k$ is the number of items to be recommended. Similarly, let $S^-$ be the set of top-$k$ negatives ordered by $s$. Then, 
\begin{equation}
\aucrelk(s) = \frac{1}{\beta k}\sum_{x_i\in S^+}\sum_{x_j\in S^-} \mathds{1} [s(x_i)\geq s(x_j)],
\label{eq:auc-rel-k}
\end{equation}
where $\mathds{1}$ is the indicator function. Note that the metric considers pairwise comparisons between top negative items and top positives items and computes how many times a relevant item has secured higher score than an irrelevant item. Thus, $\aucrelk$ rewards presence of every relevant item in top-$k$ scored elements and penalizes high scores of negative items. Also, note that the metric essentially behaves like \pauc  for $n_+\ll k$ and like \preck for $n_+\gg k$. Since, $n_+$ can be significantly different for various users (and datasets), \papk provides a more nuanced and comparative evaluation, due to which it is used as the key-performance-indicator by MSTeams' {\em production system}.

We now define the micro and macro versions of $\aucrelk$ to handle multiple users:
{\small
\begin{equation}
\text{Micro-}\aucrelk(s) = \frac{1}{P} \sum_p \aucrelk_p (s),\ \ \ \text{Macro-}\aucrelk(s)=\frac{\sum_{p}\sum_{x_i\in S^+_p}\sum_{x_j\in S^-_p} \mathds{1} [s(x_i)\geq s(x_j)]}{P\cdot k \cdot \sum_p \beta_p},
    \label{eq:microdef}
\end{equation}
}
where $\{\user_p\}_{p=1}^P$ are the set of users. We define $S^+_p$, $S^-_p$, $\beta_p$ for each user $p$ as mentioned above. Finally, $\aucrelk_p$ is $\aucrelk$ computed for user $\user_p$. In this paper, we focus on the micro-\aucrelk since it is more accommodative of varied engagement level across users and indicative of average of performance per user. For the rest of the paper, unless specified \papk refers to micro-\aucrelk. 

%% file: infer.tex
\label{ssec:inference}

% \begin{minipage}{.3\linewidth}
%     \resizebox{\textwidth}{!}{
% \begin{tabular}{@{}|l|l|l|l|@{}}
% \hline
% \textbf{Rank}  & \textbf{$f_{1}$} & \textbf{$f_{2}$} & \textbf{$f_{3}$} \\ \hline
% 1              & 1                 & 1                 & 1                 \\ \hline
% 2              & 0                 & 1                 & 0                 \\ \hline
% 3              & 1                 & 0                 & 1                 \\ \hline
% 4              & 0                 & 0                 & 1                 \\ \hline
% 5              & 1                 & 0                 & 0                 \\ \hline
% 6              & 0                 & 1                 & 0                 \\ \hline
% \textbf{AUC}   & 0.67              & 0.67              & 0.77              \\ \hline
% \textbf{pAUC(0, 2/3]}  & 0.5               & 0.67              & 0.75              \\ \hline
% \textbf{\aucrelk} & 0.75              & 1.0               & 0.75              \\ \hline
% \end{tabular}
%     }
%   \captionof{table}{Table showing a comparative example of AUC, pAUC and \aucrelk at k=2\label{metric_example}}
% \end{minipage}

\medres is trained to approximately optimize regularized \aucrelk i.e. the goal is to find parameters $\Theta$ of the \medres architecture to minimize:  $    \mathcal{L} = (1-\papk(\Theta))
    + \tau \|\Theta\|_2^2$, 
where the first term is the \papk loss function and $\tau$ is the regularizer. %and the second term is sum of $\ell 2$-norms of all the weight matrices in the architecture.  
Note that the loss function is discontinuous and hard to optimize in general. However, we propose a simple iterative procedure to refine the $\Theta$ parameter. In each iteration, we create a pool of datapoints where the pool contains the set of top-$\beta_p$ positives and top-$k$ negatives for each user $p$ according to the current scoring function. Here, $\beta_p=\min((n_+)_p, k)$ and $(n_+)_p$ is the number of positives for user $p$.  We then optimize the standard cross-entropy loss on the selected points using Adam~\cite{kingma2014adam} optimizer. We then update the score function, and again compute the set of top positives and negatives, and iterate till convergence. See Algorithm~\ref{alg:medres} for a pseudo-code of the method. Note that {\em Select-top(s,k,label)} function selects top $k$ points of given {\em label} when sorted by the score $s$. Also, the subroutine in line 8 in Algorithm~\ref{alg:medres} is fast since it runs on a few selected points. 

\begin{algorithm}[H]
\caption{Training MEDRES to  optimize \aucrelk}\label{algo}
\begin{algorithmic}[1]
\Require{TrainingData, Iterations, $k$, $P$: no. of users}
\Ensure{MEDRES model}
\Procedure{Iterative Training}{}

\State S $\gets$ TrainingData
\State model $\gets Model\_Init()$
\State $(n_+)_p=$ no. of positives for user $p$, $1\leq p\leq P$
\State $\beta_p=min((n_+)_p, k)$, $1\leq p\leq P$
\State i $\gets$ 0
\While {$i\leq\ $Iterations}

\State model.train(S)
\State {\small $scores_p  \gets$ model.score(TrainingData, p),}\par  \hskip\algorithmicindent \qquad\qquad{\small \ $1\leq p\leq P$} 
\State {\small S $\gets\cup_p$ \{ Select-top($scores_p$, k, `$-$') $\cup$} \par  \hskip\algorithmicindent \;\;\;\;\;\;\;\;\;\;\;\;\; {\small  Select-top($scores_p$, $\beta_p$, `$+$')}\}

\State i $\gets$ i+1
\EndWhile
\EndProcedure
\end{algorithmic}\label{alg:medres}
\end{algorithm}
        
\begin{figure}
\begin{minipage}{\textwidth}
	\vspace*{10pt}
\begin{minipage}{.45\textwidth}
	\resizebox{\textwidth}{!}{
		\includegraphics[scale=0.25]{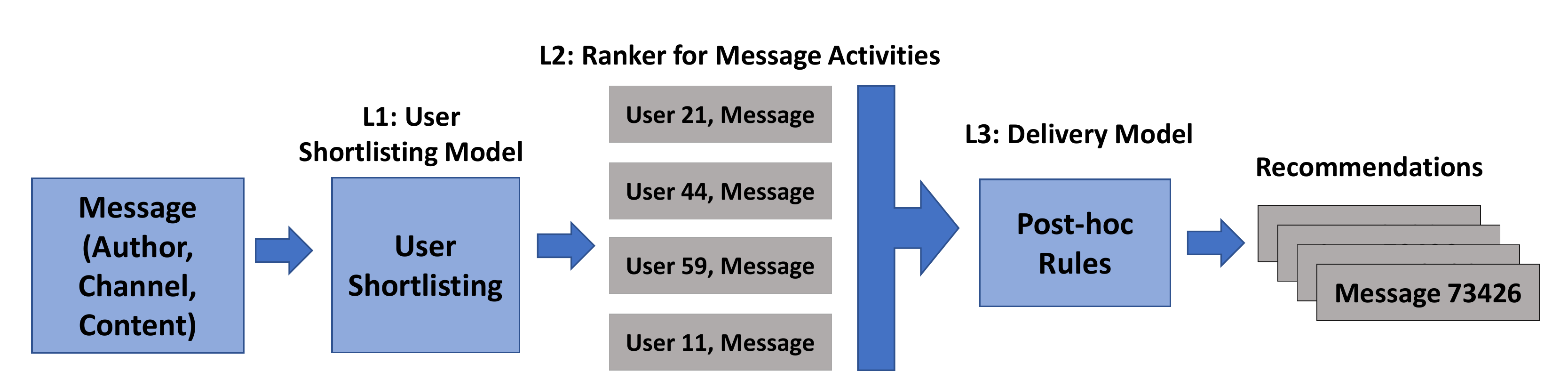}
	}
	\label{fig:prod_arch}\vspace*{-20pt}
	\captionof{figure}{The current Production System deploys a three layer architecture with layers-- L1: user weaning, L2: message classification and L3: personalized delivery model. Our model is relevant for the $L2$ layer and is evaluated against the production-grade models for the L2 layer.}
	
\end{minipage}
\begin{minipage}{0.45\textwidth}	
	\captionof{table}{Details of MS Teams Message Recommendation  Dataset\label{msgreco_dataset}}\vspace*{-10pt}
	\resizebox{\textwidth}{!}{
		\begin{tabular}{|l|l|l|l|}
			\hline
			Statistic        & Training Data & Val Data & Test Data \\ \hline
			Number of points & 3416632       & 379626   & 1499083   \\ \hline
			\% of positives  & 4.40          & 4.87     & 4.80      \\ \hline
			Users            & 901           & 898      & 898       \\ \hline
			Authors          & 677           & 509      & 593       \\ \hline
			Channels         & 476           & 367      & 391       \\ \hline
	\end{tabular}  }
\end{minipage}
		\end{minipage}
\end{figure}
% We believe that such a procedure optimizes the loss under certain restrictions, and leave further theoretical investigation into convergence and consistency of the method for future work. Note that the graph embeddings are learned only using the training data, but used in the inference time as well along with the unseen dynamic content.
During our ablation studies (Table~\ref{ablation}) we observe that the iterative procedure indeed improves \papk metric significantly over the iterations of the proposed method. 
% However, since the currently the method is not guaranteed to optimize \papk, 
We leave further theoretical investigation into convergence and consistency guarantees for future work. %Note that the graph embeddings are learned only using the training data but are used for inference along with the unseen dynamic content.

%% file: setup.tex
We now discuss the setup we use to evaluate our methods in the next three sections. We refer to the \medresM setup with \papk optimization as \medres in the further sections. 

{\bf Baselines}: For most of our experiments, we use  the standard content-filtering baseline (\conf) based on LightGBM~\cite{ke2017lightgbm} and non-linear collaborative filtering \colf as the key baselines.  We later introduce other baselines that are scenario-specific. For \conf, we first compute joint features between users and items by using one-hop graph features. That is, for each graph $G^{d,a}=(V^{d,a}, A^{d,a})$ between entity $\E^{(d)}$ of the user $\user$, and $\E^{(a)}$ of the item $\myitem$, we use edge $A^{d,a}(\user, \myitem)$ as the feature. We then collect all the features along with $\nu(\myitem)$ and $\zeta(\user)$ (see \eqref{eq:user}, \eqref{eq:item}) to form data point $\{\phi(\user_i, \myitem_i), y_i\}$.  
For \colf baseline, as is standard, we learn features for each entity directly from the given labelled data. That is, intuitively, \colf baseline is the same as \medres architecture but rather than using the graphs to regularize/extract features, it learns features from scratch to fit the given labels.

Note that we focus on \conf and \colf as baselines, because they are the most popular and widely applicable techniques used routinely in real-world systems with multiple source of information. Most of the other recommendation system techniques like GCNN-MC\cite{monti2017geometric}, HIN \cite{fu2017hin2vec} do not easily apply to the problems that we study. For example, GCNN-MC~\cite{monti2017geometric,van2017graph} in it's original form handles only user and item graphs and their labels, hence if we have other entities or a dynamic component in users/items, it will not be able to exploit that information faithfully. Similar observations hold for other methods like  inductive matrix completion \cite{jain2013provable} and matrix completion with graph side information \cite{rao2015collaborative}. 

To demonstrate the effectiveness of the \gcnp block, we also provide a comparison with the \medres-GCN architecture where the graph embedding is computed using the state-of-the-art GCN method \cite{kipf2016semi}. We also evaluate the effect of using \papk optimization in Section~\ref{sec:msgreco}. Finally, for the team recommendation problem (Section~\ref{sec:teamreco}), which is a link-prediction problem with graph side information, we use the Laplacian smoothing method by \cite{rao2015collaborative}. Furthermore, in this case, \medres-GCN reduces to the standard matrix-completion with GCN formulation \cite{monti2017geometric,kipf2016semi}. 

{\bf Metrics}: For all the benchmarks, we use \papk as the primary evaluation metric. In addition, we use \preck metric for the Teams Recommendation scenario, which is essentially a link-prediction problem, and hence \preck is a standard metric. For the remaining benchmarks and scenarios, we also compute AUC as it is the default evaluation metric for bipartite ranking style problems. 

{\bf Hyperparameters}: For \conf, we sweep on the following hyper-parameters: learning rate $\{10^{-3}, 10^{-2}\}$, number of trees $\{50, \dots, 300\}$, and number of leaves $\{10, 15, \dots, 30\}$.  For \medres-GCN, \medres and \colf, the cross-validation is done on embedding size $\{2^5, 2^6, 2^7\}$, regularization parameter $\{10^{-4}, 10^{-5}\}$, nodes in fully connected layers $\{2^7, \dots, 2^{10}\}$, batch size $\{1, 3, 5\}\times 10^3$, and learning rate $\{10^{-3}, 10^{-4}\}$. As stated earlier in the paper, we report only Micro-$\aucrelk$ (Eq. \ref{eq:microdef}) as it is more relevant metric with a large number of heterogeneous users, and refer to it as $\aucrelk$ throughout this section. For the Laplacian smoothing approach \cite{rao2015collaborative}, we tune on the following hyperparameters: learning rate $\{10^{-3}, 10^{-4}\}$, regularization parameter $\{10^{-3}, 10^{-2}, 10^{-1}\}$ and embedding size $\{2^5, 2^6\}$.

%We compare the proposed method (\medres) with a content-filtering baseline (\conf) based on LightGBM~\cite{ke2017lightgbm}, GCN within \medres \medres-GCN, and a strong end-to-end collaborative filtering baseline (\colf). The \conf baseline is similar to the production model. Also, GCN is a strong baseline since it has been known to outperform a lot of graph neural network based approaches. For the team recommendation task, we also compare \medres with a link prediction approach proposed in \cite{rao2015collaborative}. For the rest of the paper, we will denote \medresM with pAp@k optimization as \medres. %We also compare our method with a powerful end-to-end collaborative filtering (\colf) approach within the \medres framework to understand the gains of doing graph based learning by ignoring graphs in this method. %NN baseline draws the comparison with respect to using 1-hop graph features instead of static entity representations and keeping the classifier as same i.e. the fully connected layers of the architecture.

%% file: exp_teams.tex
\label{sec:msgreco}

Microsoft Teams (MSTeams) is a fast-growing enterprise conversation platform. It is composed of teams, users, and channels, where users are a part of a team, and users post on channels to start a conversation. Each user has access to all channels of a team that she/he is a part of, thus privy to a large number of messages posted per day, hence prioritizing messages is a significant challenge and creates an information overload \cite{gomez2014quantifying}. This problem can be addressed by an accurate message recommendation system that recommends the most relevant messages to users.%; MSTeams also support access control to conversation by means of one-to-one or group chats, where you need to be invited to be able to access the conversation. 

%Figure \ref{fig:TeamsSnap} shows an instance of the Teams display. The left pane shows the teams that a user is part of, along with some selected channels and middle screen shows the conversations from the ``Development'' channel.

% \begin{figure}
%   \centering
% \includegraphics[scale=0.15]{figures/Microsoft-Teams.jpg}
%   \caption{Microsoft Teams Front-end: left panel shows various teams and channels that the user is part of, while right panel shows messages from a particular channel. {\tiny Source:  \url{news.microsoft.com/apac/2017/03/15/microsoft-teams-rolls-out-to-office-365-customers-worldwide}}}
%   \label{fig:TeamsSnap}
% \end{figure}

%While communication via email are typically access controlled, i.e., you need to be explicitly part of the recipients list to get access to its content. MSTeams on the other hand enables 
%These conversations between large teams by means of access control free channels lead to a large number of messages that can be read by a user and creates significant information overload for the user, similar to social networks \cite{gomez2014quantifying}. 

%Due to these large number of possibly relevant messages, a user might ignore several of them and hence miss on some important messages which in enterprise setting can be critical. 

%This problem can be addressed by an accurate message recommendation system that recommends most relevant messages to users. %, thereby reducing the information overload. %These messages can be surfaced in the notification pan of the MSTeams product. 

The message recommendation problem can be stated as learning a function $s(\user, \myitem)$, which takes as input a user $\user$ and a message $\myitem$, and predicts whether the user would be interested in this message or not. That is, the problem is a special case of our formulation in Section~\ref{sec:prob} with $\user$ being any user of MSTeams, item $\myitem$ being any message in a channel, and label $y$ if the $\user$ {\em engages} with the message. The term {\em engagement} can be defined in various ways, here we define a user to have engaged with a message if the user likes the message or replies to the message.

Overall, the system has three entity types: $(\E^U, \E^A, \E^C)$, i.e., the users, the authors of messages, and the channels in which messages are posted. The user $\user$ is defined by only the user entity type. Message $\myitem$, on the other hand, is composed of the author of the message, channel of the message as well as the content of the message. That is, $\myitem=(\E^A_\myitem, \E^C_\myitem, \nu(\myitem))$ where $\E^A$ is the set of authors, $\E^C$ is the set of channels, and $\nu(\myitem)$ is a vector embedding of the message content. In addition, we are given 10 graphs between the three entities: $\E^U, \E^A, \E^C$. %We define several edges in these graphs to model the following relationships between users, authors and channels, as shown in Figure \ref{fig:TeamsGraphs},
\begin{enumerate}[leftmargin=2em]
    \item User-Author Graph: This is a graph between users and authors with directional edges such as the number of times a user liked an author's post, and vice versa, and number of times a user replied to author's post, etc.
    \item User-Channel Graph: This graph models user channel  interaction by capturing edges such as how many times a user visited the channel in the past month, how many messages a user posted on the channel, etc.
\end{enumerate}

{\bf Production System Architecture}: %As discussed above, the problem of message recommendation with the three entities is an instance of our formulation, and hence our learning architecture and method (Section~\ref{sec:method}) can be applied directly. However, there are some practical considerations due to which the final architecture has to be more complicated. 
MSTeams has a daily budget of $k$ messages to be recommended per user. Furthermore,  messages are recommended as they arrive, so the position of the message is irrelevant, and hence NDCG style metrics are not suitable. Instead, the problem requires bipartite ranking with limited items and highly varied engagement-per-user. Hence, \papk metric is a good fit and is used in the production system. Now, the classifier $s(\user, \myitem)$ needs to be invoked for every message $\myitem$ and the set of users who can access the message. This can be computationally infeasible due to a large volume of messages. The current production system avoids this issue by deploying a three layer architecture with layers-- L1: user weaning, L2: message classification and L3: personalized delivery model. Our model is relevant for the $L2$ layer and is evaluated against the production-grade \conf model.

\begin{minipage}{\textwidth}
  \begin{minipage}[b]{0.49\textwidth}
    \centering
    	\includegraphics[scale=0.4]{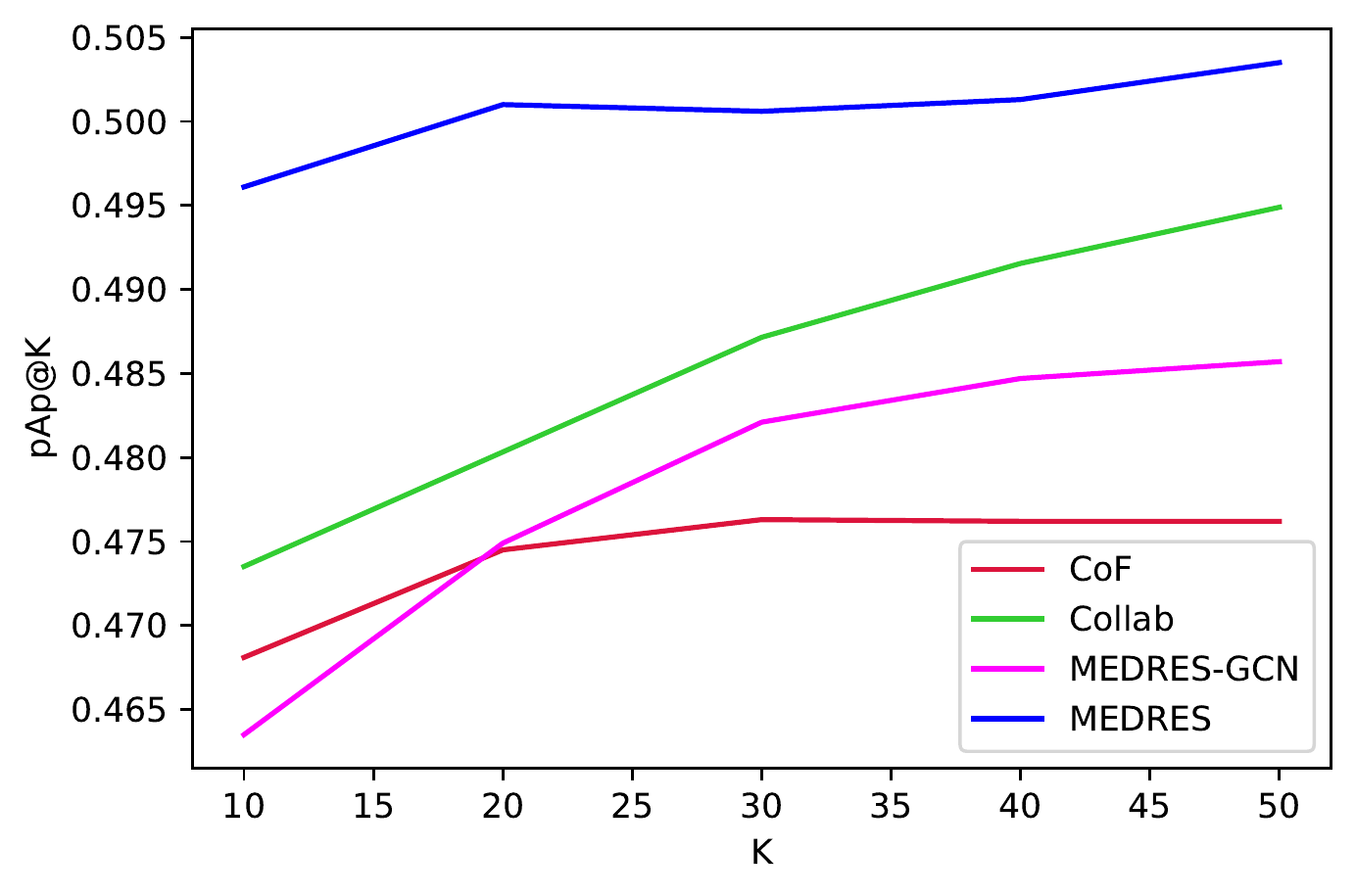}
    	\captionof{figure}{\papk accuracy obtained by various methods. \medres outperforms baselines by significant margin.}
    \label{msg_reco_result_fig}	
  \end{minipage}
  \hfill
  \begin{minipage}[b]{0.49\textwidth}
    \centering
    \begin{tabular}{|l|l|l|l|}
				\hline
				\textbf{} & \multicolumn{3}{c|}{\textbf{\#Graphs = 1}}                 \\ \hline
				k         & \conf   & {\begin{tabular}[c]{@{}l@{}}\medres-\\GCN\end{tabular}} & \medres \\ \hline
				10        & 0.447 & 0.468                   & 0.478                  \\ \hline
				20        & 0.448 & 0.474                 & 0.481                  \\ \hline
				30        & 0.452 & 0.478                  & 0.485                  \\ \hline
				40        & 0.451 & 0.478                  & 0.488                  \\ \hline
				50        & 0.452 & 0.479                  & 0.492                   \\ \hline
			\end{tabular}
			\vspace{2mm}
      \captionof{table}{Accuracy of \medres and \conf baseline with only 2 graphs}
      \label{graph_1}
    \end{minipage}
  \end{minipage}

{\bf Empirical Results}: We use a subset of the production dataset for evaluating our \medres method against various baselines (Section~\ref{sec:exp}); Table \ref{msgreco_dataset} summarizes the dataset.  Recall that we define the label of a (user, message) pair to be positive, i.e., $y=1$, if the user likes/replies to the message. Figure \ref{msg_reco_result_fig} reports ~\aucrelk \eqref{eq:microdef} accuracy for various methods. %Recall that Light-GBM model and NN model use a standard content-filtering approach where joint features for (user, message) pair are computed by concatenating standard one-hop edges from the graphs mentioned above. %For example, one such feature would be, in past, the number of times a user $\user$ liked messages from the author of message $\myitem$. 
 \medres is significantly better than both \conf and \colf baselines across various $k$ values. Furthermore, \medres with \gcnp block is also 3\% better than the \medres-GCN method. %\conf baseline is very close to the deployed production model. It can be seen that our algorithm perform significantly better than the baseline methods. For example, \medres is $\approx XX\%$ better than the \conf baselines in terms of \aucrelk with $k=30$. Relatively, we observe an increase of 4.41\% in \aucrelk. It can be further seen that \medres is approximately 1\% better than the \colf baselines and GCN. This shows that the learning the non-linear weights of different graphs helps \medres learn more relevant signals for the prediction task.
Table \ref{auc_results} shows the AUC measurements, and we observe a similar trend. \medres performs the best followed by \medres-GCN and then followed by \colf and \conf.

Next, we  evaluate \medres when only two graphs are available, instead of the 10 graphs used above. That is, 1 User-Author  graph and 1 User-Channel graph.  Figure \ref{graph_1} presents \papk accuracy for this setting. We observe that the performance of \medres and \medres-GCN is similar to the case when all 10 graphs are available. Thus, \medres is able to extract powerful features from 2 graphs while the accuracy of the production baseline (\conf) suffers. %accuracy see a significant reduction in accuracy. %It can be seen that the production baseline (\conf) significantly suffers when the number of graphs has been changed to 1, GCN baseline does not follow by the same level of degradation and is still better than \conf and \medres is significantly better than both \conf as and GCN baselines by almost 4.5\% points and 2.1\% points respectively. This shows that \medres with its non-linear weighing method helps in extraction of relevant signals even from a single graph beyond a typical GCN.

\subsection{Ablation Study}
We compare the performance of various flavors of \medres, including the impact of the \gcnp block against \medres-GCN, the impact of \papk optimization, and the impact of the aggregation function (concatenation and sum). Figure \ref{ablation} summarizes our results. Firstly, we observe that \medres has superior performance to \medres-GCN for all K values (rows 2, 3 and 6), demonstrating that the \modifiedGCN block learns powerful features.  Secondly, the proposed \papk optimization method improves accuracy for both \medres-GCN and \medres method by about $.5\%$. Thirdly, we observe that the concatenation method always performs better than the sum of functions because the summation of embeddings could lead to a loss of signal which is otherwise available in the concatenation aggregation. Finally, we also compare 1 layer and 2 layers of \modifiedGCN and show that for concatenation aggregation method, the results improve as we move to 2 layers. This shows that 2 hops of neighborhood aggregation leads to more accurate embeddings. For the sum aggregation method, we observe a different trend i.e. a decrease in performance with 2 layers because for 2 layers, the summation of embeddings is possibly leading to a massive loss of relevant signal as compared to a 1 layer \medres.

\begin{table}
\centering
\begin{tabular}{|l|l|l|l|l|l|}
					\hline {Model}      & {K=10}               & {K=20}               & {K=30}               & {K=40}               & {K=50}               \\ \hline
					{\conf (Prod Model)}                                                                             & \multicolumn{1}{r|}{0.468} & \multicolumn{1}{r|}{0.475} & \multicolumn{1}{r|}{0.476} & \multicolumn{1}{r|}{0.476} & \multicolumn{1}{r|}{0.476} \\ \hline
					{\medres-GCN Baseline} & 0.464       & 0.475  & 0.482   & 0.485        & 0.486        \\ \hline
					{\medres-GCN   + pAp@50 opt.}    & \multicolumn{1}{r|}{0.467} & \multicolumn{1}{r|}{0.481} & \multicolumn{1}{r|}{0.488} & \multicolumn{1}{r|}{0.490} & \multicolumn{1}{r|}{0.495} \\ \hline
					{\begin{tabular}[c]{@{}l@{}}\medres 1 Layer(Concatenation)\end{tabular}}   & 0.482                     & 0.491                    & 0.496                     & 0.500                   & 0.503                  \\ \hline
					{\begin{tabular}[c]{@{}l@{}}\medres 1 Layer(Sum)\end{tabular}}   & 0.476                      & 0.490  & 0.496   & 0.499                     & 0.502     \\ \hline
					{\begin{tabular}[c]{@{}l@{}}\medres 2 Layers (Concatenation)\end{tabular}}               & 0.495                    & 0.499                  & 0.499                    & 0.499                      & 0.501                      \\ \hline
					{\begin{tabular}[c]{@{}l@{}}\medres 2 Layers (Sum)\end{tabular}}                & 0.465                     & 0.475                     & 0.480                      & 0.485                      & 0.489                   \\ \hline
					{\begin{tabular}[c]{@{}l@{}}\medres 2 Layers \\ (Concatenation) + pAp@50 opt.\end{tabular}} & 0.4961    & 0.5010 &      0.501   &  0.501          &    0.504                       \\ \hline
		\end{tabular}
		\caption{Ablation results for impact of \papk optimization, aggregation method, number of layers, and \gcnp. \label{ablation}}
\end{table}

%% file: exp_teamsrec.tex
\label{sec:teamreco}

In this section, we discuss the second real-world use case of our proposed method, which is that of recommending Teams or groups (in MS Teams) that a user can join. This is a popular feature in MS Teams to increase user engagement. The problem is essentially a link prediction problem, but several users have joined limited Teams, so one needs side-information about users. This is provided in the form of a user-user graph which reflects how the user interacts with other users on certain O365 products. 

Formally, the problem is that we have two entities: Users $\E^U$, and Teams $\E^T$ and the user-item pair is also the same: Users $\user$ and Teams $\team$. The only graph is between User-User entities, given by: $\graph^{UU}$. So the problem naturally fits in the \medres architecture, although it is a significantly simpler problem from a modeling standpoint. Since it is a link prediction problem, we can also use standard techniques like KNN (which is similar to the production model), Inductive Matrix Completion (IMC) where the user features are generated using PCA of the user-user graph (referred to as PCA), the Laplacian smoothed Graph Information method by \cite{rao2015collaborative}, and finally the GCN based matrix completion technique \cite{kipf2016semi,monti2017geometric} which reduces to \medres-GCN. \conf baseline does not directly apply here as we cannot compute Teams feature from existing graphs (only graph given is $\graph^{UU}$),  and \colf baseline does not generalize to new users or users with little Team enrollment, so instead we use the PCA based IMC method mentioned above. %where the label $L$ for a (user, Team) tuple is $L = 1$ if user is enrolled in that Team and $0$ otherwise. The provided graph 

\textbf{Experiments:} We perform our experiments on a subset of the production dataset, as described in Table \ref{teams_reco_stats}.% The \conf baseline is not suitable for this experiment since \conf approach cannot be applied to this setting. The \colf baseline is also not apt for comparison here because the train-test split is based on users. If a user is not seen as part of training, \colf baseline does not learn anything for the user and in this use-case, where we are recommending teams to new users, the users in the test/validation set are the ones which are not seen in training data.   %The $\user$-$\user$ graph that we use for our analysis has 197,422 total edges, with an average node degree of 125.58.

% \begin{table*}
% \centering
% 	\caption{Teams Recommendation Dataset Statistics}
% 	\label{teams_reco_data_stats}
% \begin{tabular}{|l|l|l|l|}
% \hline
% Statistic        & Training Data & Val Data & Test Data \\ \hline
% Number of points & 8678843       & 2905245  & 2905245   \\ \hline
% Number of positives  & 33161          & 8670     & 8670      \\ \hline
% Users            & 941           & 315      & 315       \\ \hline
% Teams            & 9223          & 9223     & 9223       \\ \hline
% \end{tabular}
% \end{table*}

\begin{table}
    \centering
    	\begin{tabular}{|l|l|l|l|}
						\hline
						& Train & Val & Test \\ \hline
						No. of points & 8.7M       & 2.9M  & 2.9M   \\ \hline
						No. of positives  & 33K          & 8K     & 8K      \\ \hline
						Users            & 1K           & 300      & 300       \\ \hline
						Teams            & 9K          & 9K     & 9K       \\ \hline
						
		\end{tabular}
    \caption{Details of MS Teams Team Recommendation Dataset}
    \label{teams_reco_stats}
\end{table}

%\textbf{Metric - Precision@K:} We select \textit{Precision@K} as the metric for evaluating the quality of the generated recommendations, where K $\in$ \{1,3,5,7,9,11,13,15,25\}. 

For evaluation, we measure both \aucrelk and \preck, which are also used by the production system. %By using \preck as another metric in this case study, we were able to compare the proposed method on a known and relevant metric. 
%We use $Precision@5$ for model selection and test metrics. We also tune over the embedding dimension D $\in$ \{32,64\} and report the best results for each approach.
Fig. \ref{team_reco_results} (a) shows the \preck of the various approaches for different values of k. It can be seen that \medres outperforms all the related approaches and performs 3-4\% better than the KNN based production baseline for all $k>3$; and the performance gap increases with an increase in $k$, which is the practically relevant setting. \medres also performs 2\% better than \medres-GCN and approx. 4\% better than PCA for all $k$ values. Fig. \ref{team_reco_results} (b) shows a similar trend for \aucrelk, with the only exception being relatively poor performance of \medres-GCN as it might not be extracting strong features for relatively disconnected users on the graph; \medres with \gcnp block is the most accurate method on this dataset. % as inferior to the production model but \medres was able to significantly outperform the production model by 2-3\% across different k values.

\begin{minipage}{\textwidth}
  \begin{minipage}[b]{0.49\textwidth}
    \centering
    \includegraphics[scale=0.5]{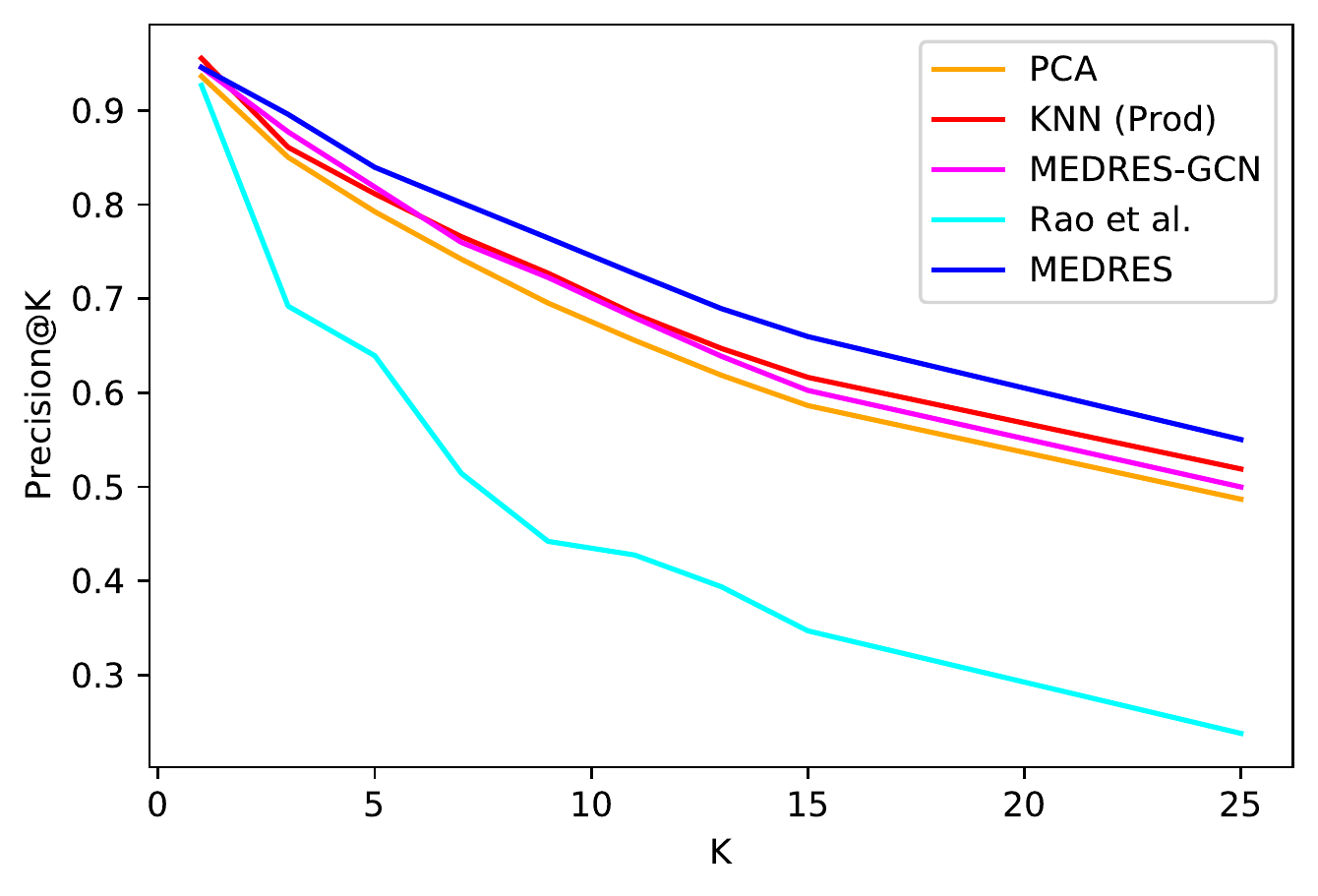}\\ (a)

  \end{minipage}
  \hfill
  \begin{minipage}[b]{0.49\textwidth}
    \centering
        \includegraphics[scale=0.5]{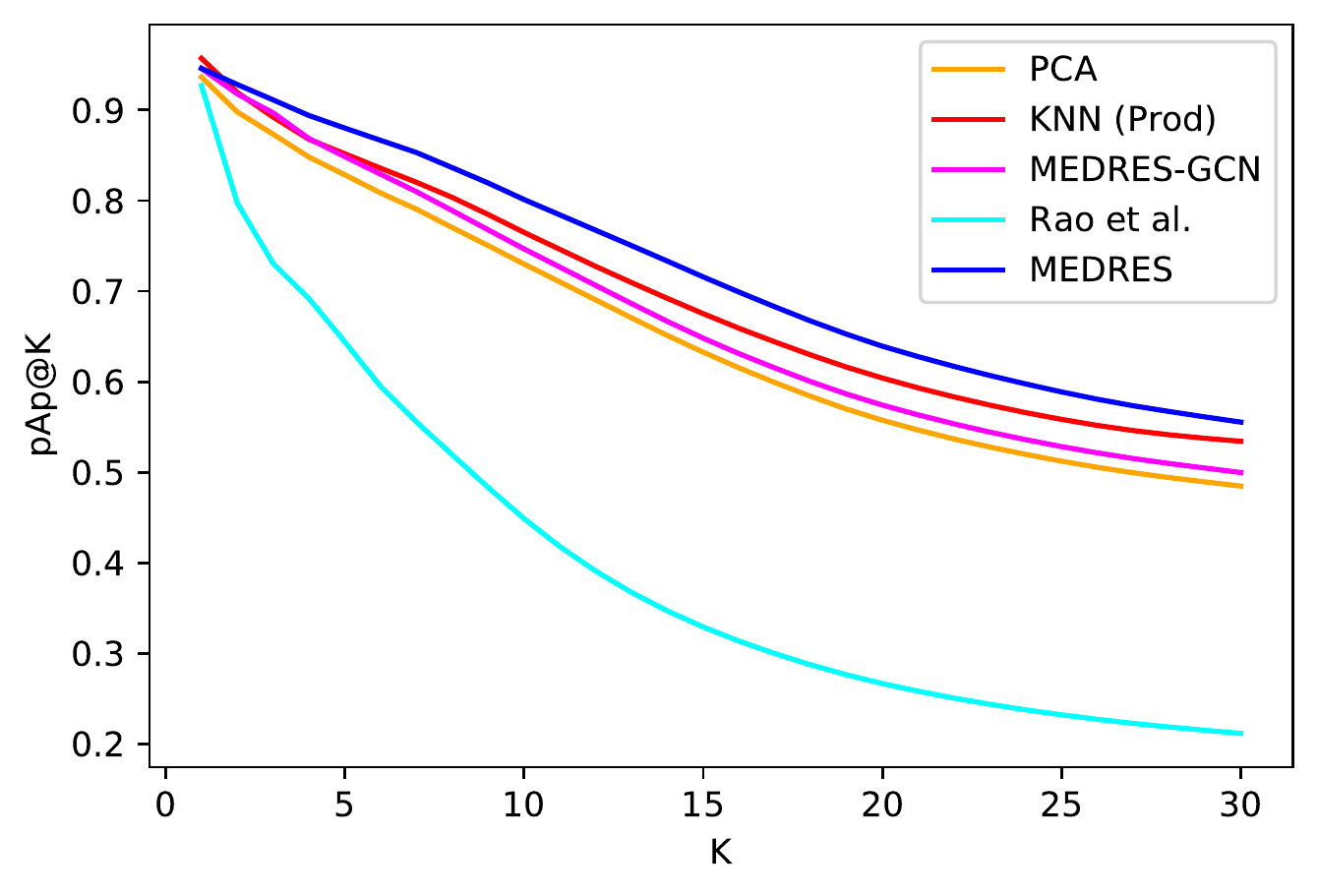}\\(b)
      %\caption{ \papk for different K values}
    \end{minipage}
\captionof{figure}{(a), (b): \preck and \papk accuracy for Teams Recommendation problem. Clearly, \medres outperforms existing baselines significantly on both the metrics. \label{team_reco_results}}
  \end{minipage}

%% file: exp_citation.tex
%\gh{Do you want to add something about reproducibility? Or mention that we run our experiments on two publicly available datasets to show wider applications of our method and in the spirit of reproducible research. Even applied/product based conferences like KDD data science track make this easy way to reject a paper these days.}
In this section, we apply our general formulation and \medres method on two publicly available datasets of citation networks and Flickr\footnote{www.flickr.com} and show the wide applicability of our proposed method. We further make the citation dataset and source code available for reproducibility along with the stated hyper-parameters in Section \ref{sec:exp}.

{\bf Citation network problem}: Here, we study the problem of  recommending relevant papers for a given research paper, i.e., recommending relevant citations for a paper. In particular, the goal is to recommend an existing paper from an author $A_j$, appearing in conference $C_j$ to a new research paper authored by $U_j$. That is, the user entity ($\user$) is a new research paper and item $\myitem$, to be recommended, is a citation. There are three entity types: user/new-author $\E^U$, authors $\E^A$, and conference $\E^C$ venues. In addition, the paper-title is taken as the dynamic component of the citation. So, $\user=(\E^U_\user, \zeta(\user))$ where  $\zeta(\user)$ is the GloVe embedding of the paper-title \cite{pennington2014glove}. Similarly, $\myitem=(\E^A_\myitem, \E^C_\myitem, \nu(\myitem))$ where $\nu(\myitem)$ is the GloVe embedding of the title of the paper-to-be-cited.  For citation networks, we consider 5 graphs: 1) User-Published-Conference graph: Number of times a user published in a conference; 2) User-Cited-Conference graph: Number of times a user cited a conference; 3) User-CoAuthor-Author graph: Number of times, a user and author were coauthors in a paper; 4) User-Cited-Author graph: Number of times a user cited an author; and 5) Author-Cited-User graph: Number of times the user (as an author) was cited by another author. We extract data from the Citation-network V1\footnote{https://aminer.org/citation} -- a publicly available citation dataset \cite{Tang:08KDD}. 
For creating entity graphs and training data, we used the  citation data from 2000-2003, while the test set was created using citation data from 2004-2005. We consider every author and citation pair for a given paper as positive data points. For generating the negative class points, we use the following method: if a user $U_i$ has cited any paper of author $A_{j}$, then all the papers of $A_{j}$ which are not cited by $U_i$ are considered as negative data points. We further sample this data such that a user should have at least 20 data points in the training set, and the author being referred should also have at least 20 data points in the training data. We further remove all rows in the test set where either the user, author, or conference did not occur in the training data. For the paper-titles, we use $50$-dimensional GloVe embedding.  Table \ref{citation_flickr_dataset_stats} provides key statistics of the training and the test dataset. 

\begin{table}
    \centering
    \begin{tabular}{|l|l|l|l|l|l|l|l|}
				\hline
				& \multicolumn{3}{l|}{Citation Dataset}                                                                    &  & \multicolumn{3}{l|}{Flickr Dataset}                                                                          \\ \hline
				& Train                              & Val                              & Test                             &  & Train                               & Val                                & Test                              \\ \hline
				No. of Points                                                        & 238k                               & 84k                              & 85k                              &  & 260k                                & 20k                                & 59k                               \\ \hline
				\% of positives                                                      & 9.99                               & 9.40                             & 9.48                             &  & 12.87                               & 12.24                              & 12.72                             \\ \hline
				\begin{tabular}[c]{@{}l@{}}Entities\\ (Count of Entity)\end{tabular} & \multicolumn{3}{l|}{\begin{tabular}[c]{@{}l@{}}Users(1602);Authors(1147);\\ Conferences(951)\end{tabular}} &  & \multicolumn{3}{l|}{\begin{tabular}[c]{@{}l@{}}Users(839);Groups(464);\\ Entities(311);Categ(1471)\end{tabular}} \\ \hline
		\end{tabular}
    \caption{Details of benchmark datasets}
    \label{citation_flickr_dataset_stats}
\end{table}

\begin{table}
    \centering
    \begin{tabular}{|l|l|l|l|l|}
					\hline
					Dataset                                                  & CoF    & Collab & \begin{tabular}[c]{@{}l@{}}MEDRES\\ GCN\end{tabular} & MEDRES \\ \hline
					\begin{tabular}[c]{@{}l@{}}Message \\ Reco.\end{tabular} & 0.8901 & 0.8973 & 0.9026                                               & 0.9029 \\ \hline
					\begin{tabular}[c]{@{}l@{}}Citation\\ Reco.\end{tabular} & 0.7645 & 0.8198 & 0.8105                                               & 0.8156 \\ \hline
					\begin{tabular}[c]{@{}l@{}}Group\\ Reco.\end{tabular}    & 0.7535 & 0.7404 & 0.7651                                               & 0.7874 \\ \hline
		\end{tabular}
    \caption{AUC ROC Results}
    \label{auc_results}
\end{table}

{\bf Flickr dataset}: Here, we consider the task of recommending relevant Flickr groups for users to post their images. Flickr has various groups based on different themes where each group can be associated with several categories as well as tags. The original dataset denotes tags as entities, but we avoid that term to avoid confusion. We want to recommend groups to photos, so we refer to a particular group as item $\myitem$ and a photo as a user/document $\user$. $\user$ is represented by the specific user entity $\E^{(U)}$ and the content of the photograph itself, i.e., $\user=(\E^{(U)}, \zeta(\user))$; we embed images into 4096 dimensional features using VGG19 ~\cite{simonyan2014very} and then do 100 dimensional PCA reduction of the same to consume as image features. Each item is just the group entity itself, i.e., $\myitem=(\E^{(G)})$. In addition, we have the following engagement graphs: 1) User-PublishedIn-Group: Number of posts in a group by a user; 2) User-PublishedIn-Category: Number of posts by a user in a category; and 3) User-PublishedIn-Entity: Number of posts by a user in groups containing the entity type under consideration. We use the publicly available Flickr dataset ~\cite{wang2017flickr}. For each photo posted by a user in a particular group, we create a positive class data point. We  further create  negative class data points for a user and photo pair from all groups which belong to the same category as the original group of the photo. We then randomly sample the dataset to consider only 10\%  of the data-points for generating train and test datasets. Table \ref{citation_flickr_dataset_stats} provides statistics for the dataset.

\begin{figure}
	\begin{tabular}{cccc}\hspace*{-8pt}
		\includegraphics[width=.49\textwidth]{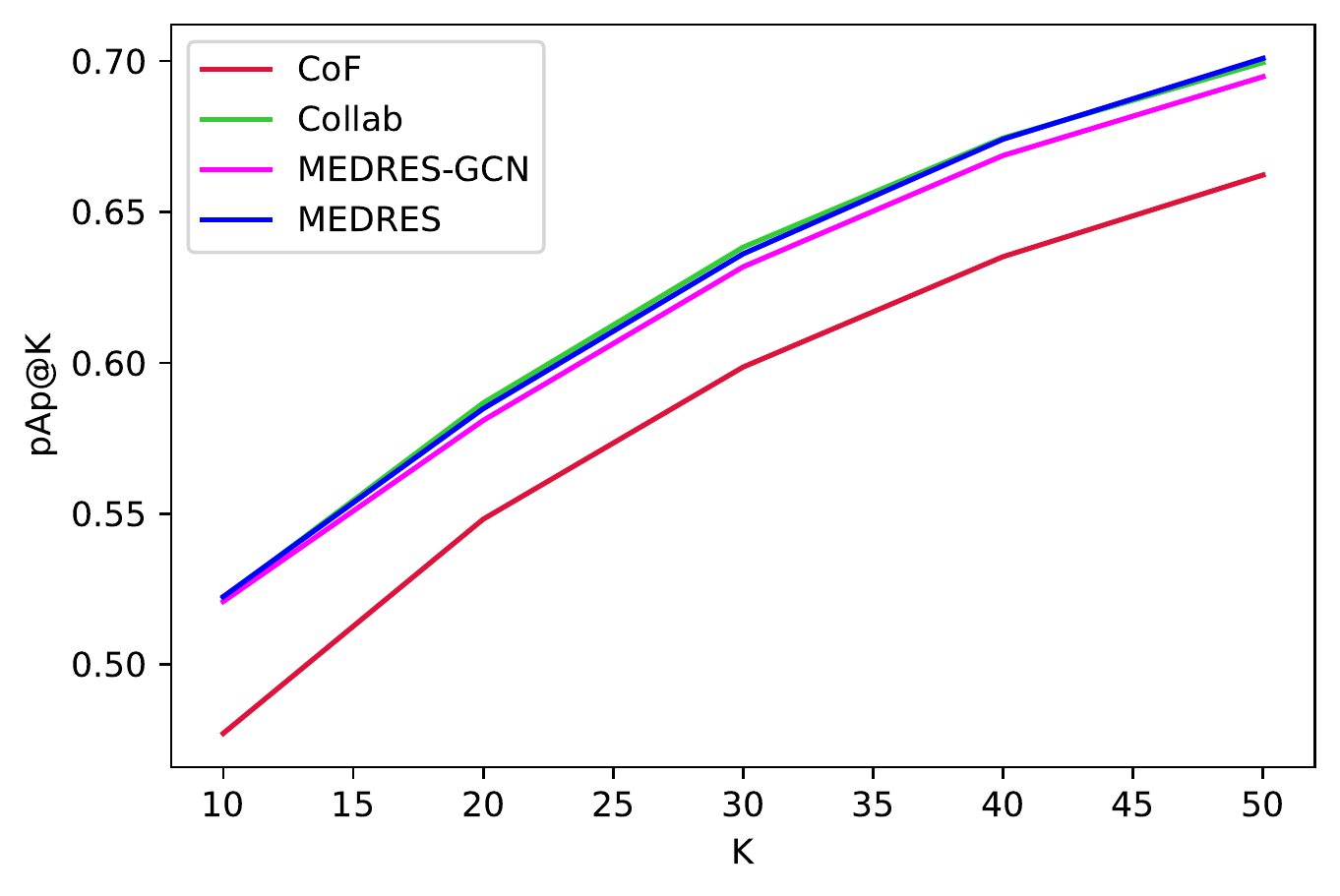}&\hspace*{-12pt}
		\includegraphics[width=.49\textwidth]{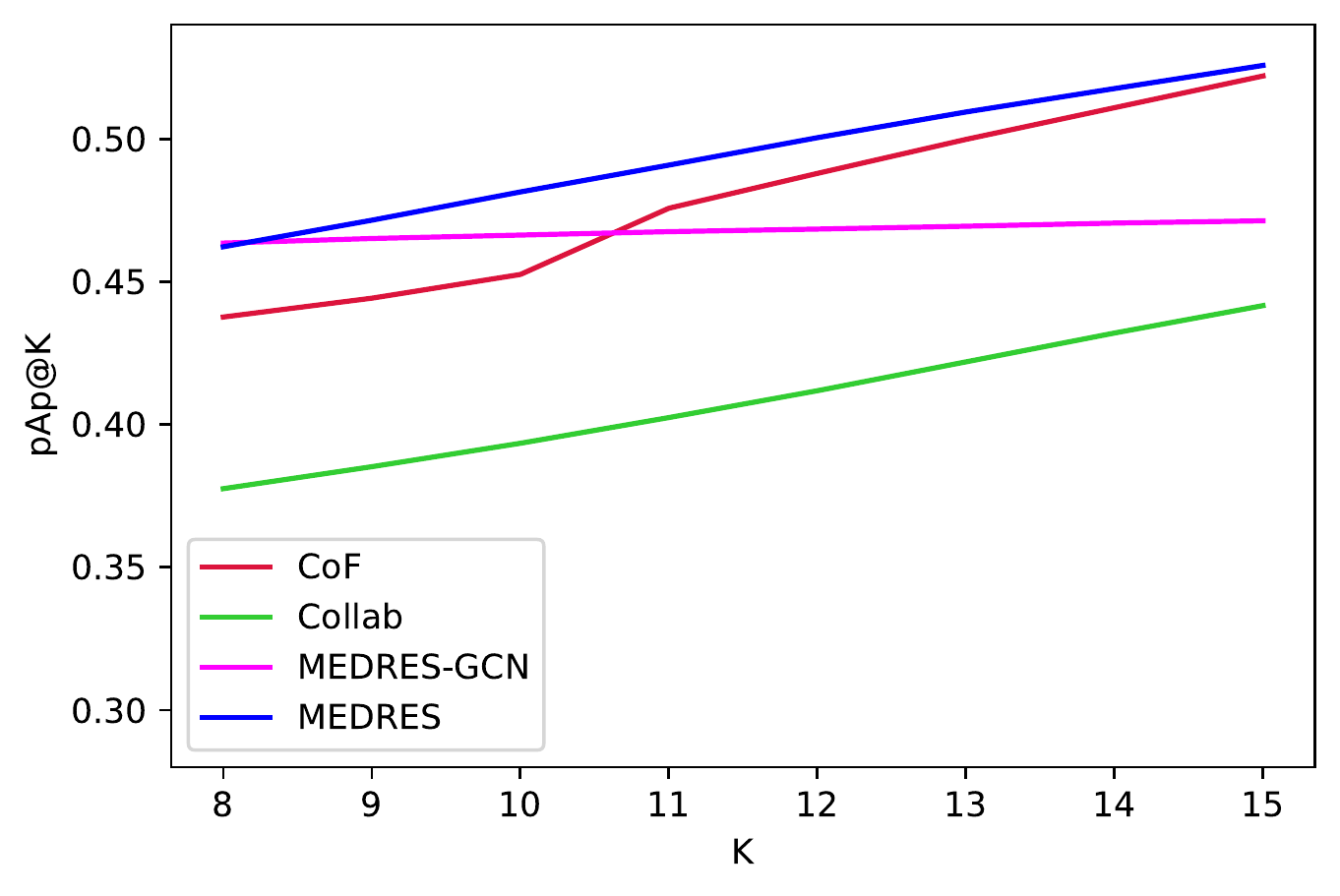}\\[-5pt]
		{(a)}&{(b}\vspace*{-12pt}
	\end{tabular}
	\caption{(a): \papk accuracy on Citation networks dataset, (b): \papk accuracy on Flickr dataset. For Flickr, \medres is 4-5\% more accurate than baselines.}\label{fig:experiments_4}
\end{figure}

{\bf Empirical Results:} We applied \medres with the same hyperparameters, as discussed in the experimental setting section.
Figure~\ref{fig:experiments_4}(c) shows \aucrelk obtained by various methods with varying $k$ for the citation recommendation problem.  For recommending $k=10$ articles, \medres-GCN is $(\geq) 4.37$\% more accurate than the \conf baselines, and similar trend holds for varying values of $k$, showing that \medres, even with vanilla GCN can outperform content filtering model. We further observe an increase of 0.5\% increase for \aucrelk when using \modifiedGCN over GCN and \colf. To understand the gain of \medres over \colf, we computed results when only 50\% data is used in training. While for 100\% data the performance of \colf was close to \medres, for 50\% data, we saw a drop in \aucrelk of 3.2\% \colf compared to only a 1.5\% drop in \medres, with pAp@50 for \colf at 0.6658 and while for \medres it is 0.6870. Table \ref{auc_results} shows the AUC values where we observe that the \colf baseline has a slightly higher value than \medres. This results from the issue that different users have different numbers of data points as stated in Section \ref{sec:papk}. To further understand it, we computed the average of per user AUC (micro-AUC). For \medres, its value is 0.7427 and for the \colf baseline, it is 0.7426 which is a reverse trend compared to AUC. This clearly indicates need for more fine-grained metric like \papk. Moreover, the \aucrelk trend remains clear across datasets.

Finally, we observe a similar trend for the Flickr group problem (Figure~\ref{fig:experiments_4}(d)) where \medres performs 2.5\% better than \medres-GCN, 3\% better than \conf and approximately 9\% better than \colf at k = 10. Thus demonstrating that \medres can extract signals which are otherwise not available in \conf and \colf baselines.

%\pj{Can we say that we will provide dataset for citation and flickr, and code in supplementary material?}

%% file: conclusion.tex
We considered the problem of constructing a recommendation system with "rich" users/items, with multiple entities expressed through multiple engagement/relationship graphs. We showed that our framework encapsulates several existing formulations, and used it to develop a novel \medres architecture that pools together signals from the graphs via Grah Neural Networks (GNN). We proposed \modifiedGCN, a novel GNN block that can be used with \medres architecture to learn graph weights and extract more powerful task-specific signals. Our training method can optimize \medres for the proposed \papk metric. Finally, we used our framework to model several different recommendation problems, including two real-world tasks, and demonstrated effectiveness of \medres against different baselines.  

Due to the generality of  our framework, it can be applied with several other embedding techniques like Heterogeneous Information Networks (HIN) that can be richer in some settings than GCN and \gcnp. Furthermore, further exploring optimization of \papk metric and studying it's theoretical properties is also an exciting research direction. 
%We believe that the technique can be of use in a variety of domains, so the application and generalization of the approach to several other interesting recommendation systems is an exciting direction for future works. Our newly proposed \aucrelk metric captures nuances of various applications, especially the MSTeams application and the current optimization strategy improves \aucrelk. As a part of future work, we believe that studying its theoretical properties is also an exciting research direction. 